%% file: main.tex
\documentclass[lettersize,journal]{IEEEtran}
\usepackage{amsmath,amsfonts}
\usepackage[linesnumbered,ruled,vlined]{algorithm2e}
\usepackage{array}
\usepackage[caption=false,font=normalsize,labelfont=sf,textfont=sf]{subfig}
\usepackage{textcomp}
\usepackage{stfloats}
\usepackage{url}
\usepackage{verbatim}
\usepackage{graphicx}
\usepackage{xcolor}
\usepackage{cite}
\usepackage{multirow}
\usepackage{hyperref}
\usepackage{colortbl}
\usepackage{enumitem}
\usepackage{tikz} 
\usetikzlibrary{external} 
\newcommand{\better}[1]{\textcolor[HTML]{05b90c}{#1}}
\newcommand{\worse}[1]{\textcolor{red}{#1}}

\usepackage[most]{tcolorbox}
\newtcbox{\inlinebox}[1][]{
    on line,
    colback=gray!25,
    colframe=gray!25, %
    boxrule=0pt, %
    arc=2pt, %
    boxsep=2pt, %
    left=0pt, right=0pt, top=0pt, bottom=0pt, %
    #1
}

\newtcbox{\highest}[1][]{
    on line,
    colback={rgb,1:red,0.66;green,0.82;blue,0.56}, %
    colframe={rgb,1:red,0.66;green,0.82;blue,0.56},
    boxrule=0pt,
    arc=2pt,
    boxsep=2pt,
    left=0pt, right=0pt, top=0pt, bottom=0pt,
    #1
}

\newtcbox{\lowest}[1][]{
    on line,
    colback={rgb,1:red,0.93;green,0.58;blue,0.58}, %
    colframe={rgb,1:red,0.93;green,0.58;blue,0.58},
    boxrule=0pt,
    arc=2pt,
    boxsep=2pt,
    left=0pt, right=0pt, top=0pt, bottom=0pt,
    #1
}

\SetKwInput{KwRequired}{Required}

\begin{document}

\title{Vec2Face+ for Face Dataset Generation} %

\author{Haiyu Wu,~\IEEEmembership{Student Member,~IEEE}, Jaskirat Singh, Sicong Tian~\IEEEmembership{Student Member,~IEEE}, \\Liang Zheng~\IEEEmembership{Senior Member,~IEEE}, Kevin W. Bowyer~\IEEEmembership{Fellow,~IEEE}
\thanks{Haiyu Wu and Kevin W. Bowyer are with the Department of Computer Science and Engineering, University of Notre Dame.}
\thanks{Jaskirat Singh and Liang Zheng are with the School of Computing, Australian National University.}
\thanks{Sicong Tian is with the School of Natural Sciences, Indiana University South Bend.}
}

\maketitle

\begin{abstract}
When synthesizing identities as face recognition training data, 
it is generally believed that large inter-class separability and intra-class attribute variation are essential for synthesizing a quality dataset. %
This belief is generally correct, and this is what we aim for. However, when increasing intra-class variation, existing methods overlook the necessity of 
maintaining intra-class identity consistency. %
To address this and generate high-quality face training data, we propose Vec2Face+, a generative model that creates images directly from image features and allows for continuous and easy control of face identities and attributes.   
Using Vec2Face+, we obtain datasets with proper inter-class separability and intra-class variation and identity consistency using three strategies: 1) we sample vectors sufficiently different from others to generate well-separated identities; 2) we propose an AttrOP algorithm for increasing general attribute variations; 3) we propose LoRA-based pose control for generating images with profile head poses, which is more efficient and identity-preserving than AttrOP. %
Our system generates VFace10K, a synthetic face dataset with 10K identities, which allows an FR model to achieve state-of-the-art accuracy on seven real-world test sets. Scaling the size to 4M and 12M images, the corresponding VFace100K and VFace300K datasets yield higher accuracy than the real-world training dataset, CASIA-WebFace, on five real-world test sets.  This is the first time a synthetic dataset beats the CASIA-WebFace in average accuracy. In addition, we find that only 1 out of 11 synthetic datasets outperforms random guessing (\emph{i.e., 50\%}) in twin verification and that models trained with synthetic identities are more biased than those trained with real identities. Both are important aspects for future investigation. 
Code is available at \textcolor{blue}{\url{https://github.com/HaiyuWu/Vec2Face_plus}}%
\end{abstract}


\section{Introduction}

Synthesizing face images to enable large-scale training sets for FR models is the way to address privacy
concerns over web-scraped datasets of real face images~\cite{competition-1, competition-2, competition-3, kim202550}. 
It is generally recognized that a good training set for FR should have high inter-class separability \cite{dcface, idiff23, vec2face25} and large intra-class attribute variation \cite{synface, sface22, dcface, idiff23, vec2face25}.

We generally agree on the importance of inter-class separability and intra-class variation. Previously we designed Vec2Face \cite{vec2face25} which excels at generating separable identities. By generating face images from sampled vectors, Vec2Face greatly increasing the capacity to over 300K identities. %

In this article, we aim to improve Vec2Face in terms of its intra-class properties. Existing works report that intra-class attribute variation affects FR accuracy~\cite{skin-tone, brightness-effect, pie,multi-pie, pie-pami,tiny-face,xqlfw,beard-area-effect1, beard-area-effect2, beard-area-effect3, comprehensive-attr, bald-effect, eye-glasses, ijbc, tian2024impact, mustache-effect, comprehensive-attr}. 
But it remains unknown how intra-class attribute variation contributes to the accuracy gap between synthetic and real training data. In Fig. \ref{fig:intra-attr}, we compare the distributions of 9 attributes across five synthetic training sets and CASIA-WebFace~\cite{casia-webface} to directly analyze  intra-class attribute variation. Interestingly, existing synthetic datasets have an on-par and sometimes even higher intra-class attribute variation than CASIA-WebFace, suggesting that neglected factors may be a cause of the accuracy gap.

\input{figures/teaser}

We argue that intra-class identity consistency is the missing puzzle. 
In Fig. \ref{fig:intra-consis} and Fig. \ref{fig:intra-inconsis}, we analyze the intra-class identity consistency across these training sets, and 
find that synthetic training sets are inferior on this dimension. In other words, synthetic images that are supposed to be of the same identity may turn out to be of diffferent identities. This would introduce noise in training. 

\input{figures/vec2face-paradigm}
To synthesize effective facial training data, this article builds on Vec2Face and extend it to Vec2Face+, a holistic approach that achieve proper inter-class separability, intra-class variation, and identity consistency. 
Vec2Face+ learns image generation by understanding the vectors extracted by a feature model (\emph{i.e.}, an FR model), see Fig.~\ref{fig:paradigm}, such that the correlation of vectors can be accurately reflected in the image domain. 
In inference, the separability of the generated identities can be effectively controlled by restricting the cosine similarity (\emph{i.e.} $\leq$ 0.3) between sampled vectors. 
Different from Vec2Face, Vec2Face+ achieves better identity consistency in the generated datasets and is more efficient to train. First, Vec2Face achieves attribute control using a gradient descent algorithm. Its iterative generation manner is time-consuming and can compromise the identity in the face image when the target attribute is extreme, \emph{e.g.}, pose $>70^\circ$. Vec2Face+ handles pose control differently, by injecting pose conditions given by face landmark points into LoRA~\cite{lora} fine-tuning. This enables a one-time generation of images with large  pose variation while preserving identity. Second, we observe that the GAN loss in \cite{vec2face25} only incrementally improves image quality while incurring substantial computing overhead. Thus, the GAN loss is dropped from the total loss, resulting in a 20\% reduction in training time.

Vec2Face+ allows us to generate VFace10K, a synthetic training set of 10k identities. FR models trained on VFace10K achieve very competitive average accuracy 93.89\% on five real-world test sets. This is 1.59\% higher than the second-best synthetic training set of the same size. Vec2Face+ also makes it easy to scale up the training set by sampling more feature vectors separated from existing ones. When we sample 100k and 300k identities, \emph{i.e.,} the VFace100K and VFace300K datasets, we report further improved accuracy of 94.88\% and 94.93\%, respectively. Both numbers are higher than that from CASIA-WebFace (94.79\%), marking the first time training on a synthetic dataset surpasses the average accuracy of a real-world dataset on the five standard test sets Fig.~\ref{fig:teaser}. 

To further evaluate  dataset quality, we include 8 more test sets covering facial attributes variation~\cite{hadrian-eclipse, ijbb, ijbc}, similar-looking persons challenge~\cite{Doppelver, hadrian-eclipse}, and demographic bias~\cite{bfw, ba-test}.  Important observations are as follows. 1) Comparing with a larger-scale dataset with real identities, WebFace4M~\cite{webface260m}, the accuracy gap is still large; 2) Regardless of the dataset scale, only 1 out of 11 synthetic training sets can achieve an accuracy slightly better than a random guess (\emph{i.e.}, 50\%) on an identical twins test set; 3) The FR model trained with a synthetic dataset has a larger demographic disparity on accuracy than that trained with a real dataset. This points to two key challenges for future research: improving twins verification and reducing demographic bias. 
The main contributions are summarized below.

\begin{itemize}[leftmargin=*]
    \item We introduce Vec2Face+, an improved approach to generating synthetic images sets for training face matchers, which achieves the first instance of a synthetic training set resulting in a face matcher with better accuracy than a widely-used, same-size real-image training set.
    \item Our experimental analysis points to intra-class identity consistency as the main factor holding back the accuracy achieved by current approaches to generating synthetic image sets for training face matchers.
    \item This paper also reveals that the similarity-based definition of identity used in current methods of generating synthetic image sets as causing matchers trained on synthetic images to fail on the task of discriminating between identical twins.
\end{itemize}

\section{Related Work}
\textbf{Discrete class label conditioned image generation.}
VQGAN\cite{vqgan}, MAGE\cite{mage}, DiT\cite{dit}, U‑ViT\cite{uvit}, MDT\cite{mdt}, and VAR\cite{var} rely on discrete class labels to control the object categories in their outputs. Because they condition on fixed label sets, they cannot create unseen classes and are unsuitable for large‑scale face‑identity synthesis. In contrast, Vec2Face~\cite{vec2face25} and Vec2Face+ condition on continuous face embeddings, allowing them to generate an unlimited number of novel identities through a suitable feature sampling.

\textbf{Identity feature conditioned image generation.}
One strategy is to feed a fixed identity vector to the generator so it produces images of a single person. FastComposer~\cite{fastcomposer}, PhotoVerse~\cite{photoverse}, and PhotoMaker~\cite{photomaker} obtain this vector with CLIP\cite{CLIP}, but their results inherit CLIP’s weak facial encoding capability\cite{arc2face}. To improve identity fidelity, InstantID~\cite{instantid}, Face0~\cite{face0}, IP‑Adaptor~\cite{ip-adaptor}, FaceRendering~\cite{facerendering}, and Arc2Face~\cite{arc2face} instead take identity features extracted by a FR model.
However, these approaches still require an auxiliary module (\emph{e.g.}, ControlNet\cite{controlnet}) to create images with diverse face attributes. This approach struggles with abstract attributes, such as age.
Different from these designs, Vec2Face and Vec2Face+ use dynamically perturbed identity features that remain highly similar to the original embedding, thereby maintaining strong identity consistency while enabling controlled attribute variation. Experimental results indicate that this approach helps the model learn a good representation for face aging.

\textbf{Synthetic face image datasets.}
Existing methods for synthetic face dataset generation fall into two categories, GAN‑based and diffusion‑based.  Each has notable limitations.
Among GAN‑based techniques, SynFace~\cite{synface}, Usynthface~\cite{usynthface}, SFace~\cite{sface22}, SFace2~\cite{sface224}, and ExfaceGAN~\cite{exfacegan23} rely on pre‑trained GAN generators that are not optimized in an identity-aware manner. In contrast, our method incorporates face features explicitly during training, enabling identity‑aware synthesis from the outset.

Diffusion‑based solutions adopt different strategies but have similar shortcomings. DCFace~\cite{dcface} couples a pre‑trained diffusion model~\cite{ilvr} with an auxiliary style‑transfer network to enhance attribute diversity and inter‑class identity separability, but this pre-trained model is not identity-aware. IDiff‑Face~\cite{idiff23} uses the identity embedding as a condition on the latent diffusion model, but the pre-trained encoder and decoder of VQGAN~\cite{vqgan} compromise the identity guidance. Arc2Face~\cite{arc2face} fine‑tunes a stable diffusion model~\cite{SD} on a high-quality version of WebFace42M to improve the generalizability on face generation. It combines the identity features and CLIP\cite{CLIP} embeddings to preserve the control of the identity of the output images. A ControlNet~\cite{controlnet} is used to expand pose variation.
However, neither stable diffusion nor CLIP was designed for facial data, and the image generation speed strongly limits the scalability of the dataset. Conversely, our model is specifically designed for face dataset generation that is identity-aware and efficient in dataset generation. This allows us to easily scale the dataset size to over 10M images and achieve state-of-the-art FR accuracy.

\input{figures/facial-attr}
\section{Dataset attributes analysis}
\label{sec:attr-analysis}

To investigate the causes of the accuracy gap, we analyze five synthetic datasets and one dataset with real identities in three aspects: 1) intra-class attribute variation, 2) inter-class identity separability, and 3) intra-class identity consistency. The datasets involved are DigiFace (Digi.)~\cite{digiface-1m}, SFace (S.)~\cite{sface22}, IDiff-Face (IDiff.)~\cite{idiff23}, DCFace (DC.)~\cite{dcface}, HSFace10K (HS.)~\cite{vec2face25}, and CASIA-WebFace (CAS.)~\cite{casia-webface}. The HSFace10K is generated by Vec2Face~\cite{vec2face25}.

\subsection{Intra-class attribute variation analysis}
\label{sec:intra}
Prior works~\cite{synface, digiface-1m, dcface, sface22, sface224, idiff23, arc2face, exfacegan23, idnet23, cemiface, hyperface, langevin-disco} simply report the accuracy of the trained face recognition model on age~\cite{agedb-30, calfw} and pose~\cite{cfpfp, cplfw} test sets to indirectly indicate the robustness of variation in facial attributes of the created datasets.
The focus on only age and pose limits the research horizon on other attributes. Hence, we directly obtain the attribute values of nine facial attributes that are known to have an important effect on face recognition accuracy to systematically analyze the effect of intra-class attribute variation. These attributes include face brightness~\cite{skin-tone, brightness-effect, pie,multi-pie, pie-pami}, face image quality~\cite{tiny-face,xqlfw}, facial expression~\cite{pie, multi-pie, pie-pami}, beard area~\cite{beard-area-effect1, beard-area-effect2, beard-area-effect3, comprehensive-attr}, baldness~\cite{bald-effect}, eyeglasses~\cite{eye-glasses, ijbc, tian2024impact}, and mustache~\cite{mustache-effect, comprehensive-attr}. 

\textbf{Obtaining attribute labels.} We obtain face brightness values using BiSeNet\footnote{https://github.com/zllrunning/face-parsing.PyTorch}~\cite{bisenet} and the Face Skin Brightness metric~\cite{brightness-effect}. Images with a segmented skin area less than 20\% of the image size are discarded to ensure accurate brightness measurement. This results in dropping less than 7\% of images. The quality of face images is measured by the magnitude of the feature extracted by the pre-trained MagFace~\cite{magface}. Age values are predicted by a classifier in~\cite{age-model} that predicts discrete age labels from 0 to 100. Head pose is described by pitch, yaw, and roll. Nevertheless, roll variation is reduced in the face cropping and alignment, and pitch variation is generally small. Thus, we  focus on yaw angle, obtained by using img2pose~\cite{img2pose}. Facial expression is also known to effect recognition accuracy, so we use  DDAMFN~\cite{expression} trained on the Affectnet~\cite{affectnet} to obtain expression labels.

Besides the aforementioned well-known facial attributes that effect recognition accuracy, we extend the breadth by adding four more attributes. Beard area, baldness, and mustache attribute labels are obtained from LogicNet~\cite{logicnet} trained with FH41K~\cite{beard-area-effect1,logicnet}. Due to these attributes being highly correlated to males, we filter out the female images using FairFace~\cite{fairface} predictions before analysis. The last attribute is eyeglasses;
each image is categorized as no eyeglasses, eyeglasses, or sunglasses. The eyeglasses classifier proposed in~\cite{tian2024impact} is used.

\textbf{Attribute variation analysis.}
We visualize the attribute variation in Figure~\ref{fig:intra-attr}. For simplicity, we use the average of intra-class attribute mean and standard deviation (std.) values to present the variation of face brightness, face quality, age, and yaw angle. As for the other five attributes, we use the average values of the attribute fractions %
to perform the analysis. The main observation from these results is that the 
robustness of facial attribute variation in the synthetic face datasets is overall on par with that in the dataset of images of real persons.

Specifically, for the five well-known attributes, HSFace10K has the largest variation in face quality; The images in HSFace10K have the best exposure level~\cite{brightness-effect}; HSFace10K has the largest variation in yaw angle; HSFace10K and IDiffFace surpass the other datasets in age variation; DigiFace has the best  representation of all facial expressions. 
For the other four attributes, these datasets have a similar intra-class variation. This indicates that existing methods of creating synthetic datasets are capable of generating variation in facial attributes that is on par with that seen in existing datasets of real face images.

Interestingly, based on the accuracy reported in Table~\ref{tab:performance}, we notice that the datasets with %
the largest variation in each attribute do not achieve the best accuracy on the test sets focused on that attribute. In other words, a large intra-class variation in a facial attribute does not guarantee good accuracy on that attribute. For example, CASIA-WebFace has the second smallest variation in yaw angle but achieves the best accuracy on CFP-FP. 
This suggests that there is another factor limiting the accuracy achieved using synthetic image training sets.
This observation motivates us to analyze another widely reported factor, inter-class separability.

\input{figures/inter-sep}
\subsection{Inter-class separability analysis}
\label{sec:inter-calc}
Inter-class separability is 
typically measured as the fraction of identities that have cosine similarity against all other identities below a given threshold value \cite{dcface, vec2face25}, so-called ``well-separated identities.''
We use a pre-trained FR model to obtain the image features and calculate the average feature of each identity to get the identity's features. Following~\cite{vec2face25}, we use 0.4 to calculate the number of well-separated identities $\mathcal{N}_{sep}$ for each dataset. The fraction of well-separated identities is then:
\begin{equation}
    \mathcal{D}_{sep} = \frac{\mathcal{N}_{sep}}{\mathcal{N}_{total}}
\end{equation}
where $\mathcal{N}_{total}$ is the number of identity folders in the dataset. Fig.~\ref{fig:inter-sep} shows that the HSFace10K created by Vec2Face  achieves comparable inter-class identity separability to CASIA-WebFace. 
However, despite achieving a fraction of well-separated identities on par with the real-image training set, the synthetic training set still has a large gap in average accuracy (\emph{i.e.}, 2.79\%) compared to real-image training set.
This suggests that there is still some other factor holding back the accuracy achieved with the synthetic training set.

\input{figures/inconsistency}
\input{figures/intra-consis}

\subsection{Intra-class identity consistency analysis}
\label{sec:intra-calc}
Browsing the images inside a given identity folder, as illustrated in Fig.~\ref{fig:intra-inconsis}, 
quickly brings the realization that identity is not well preserved in synthetic datasets.
There may even be visually apparent differences in gender and race for the images in one identity folder. 
This suggests that intra-class identity consistency may be the neglected factor.
To verify this, we analyze the intra-class identity consistency of the datasets based on cosine similarity. The equation for intra-class identity consistency measurement is:
\begin{equation}
    \mathcal{D}_{consis} = \frac{1}{\mathcal{N}\mathcal{K}}\sum_{i=1}^{\mathcal{N}}\sum_{j=1}^{\mathcal{K}}\frac{f_{i} \cdot f_{j}}{||f_{i}||\ ||f_{j}||},
\end{equation}
where $\mathcal{N}$ is the number of identities, $\mathcal{K}$ is the number of images for an identity, and $f$ is the image feature vector extracted by a pre-trained FR model. Fig.~\ref{fig:intra-consis} shows that all the synthetic datasets have lower identity consistency than the real dataset (CASIA-WebFace), and most are markedly lower.

Overall, our analyses can be summarized into three points: 1) intra-class attribute variation is well-addressed by the existing generative methods, 2) good inter-class separability can be achieved by the Vec2Face generation paradigm, but 3) it is currently challenging to combine both strong intra-class attribute variation and strong intra-class identity consistency.
\section{Methods and VFaces}
To address the challenges outlined above, 
we propose the Vec2Face+ model. 
Vec2Face+ builds on the strength of Vec2Face~\cite{vec2face25} in inter-class separability by adding an improved ability to maintain consistency of intra-class identity and also by making the training more computationally efficient. We combine the Vec2Face+ model with three dataset generation strategies to propose four synthetic training sets with 10K, 20K, 100K, and 300K identities, to compare with the datasets with real identities at different scales.

\subsection{Vec2Face+: Architecture and Loss functions}
Vec2Face+ has two components: a main model and a pose control model. 
The main model differs from the Vec2Face architecture by not using the patch-based discriminator during training. 
This reduces training time by 20\% without degrading image quality. The pose control model, additionally, includes a simple CNN block and LoRA parameters. It injects the pose condition given by face landmark points into LoRA fine-tuning, resulting in a more efficient image generation with large pose variation and mitigating the identity degradation that occurs when using the AttrOP algorithm.
The main model, as shown in Fig.~\ref{fig:vec2face-arch}, consists of a feature expansion layer, a feature masked autoencoder layer, and an image decoding layer.

\input{figures/vec2face-training}
\textbf{Feature vector preparation.} 
The essence of the Vec2Face+ design is that feature vectors should encapsulate both identity and face attributes. 
To this end, we use a pre-trained FR model~\cite{arcface} to obtain the image features.

\textbf{Feature expansion.} The extracted features are expanded from ($\mathcal{N}$, 512) to ($\mathcal{N}$, 49, 768), where $\mathcal{N}$ is the number of samples in a batch. There are two reasons for this design: 1) This 2-D feature map well matches the ViT-B~\cite{vit} input shape; 2) Combined with a 4-layer image decoder, the output image has the same shape (\emph{i.e.}, 112$\times$112$\times$3) as the original images.

\textbf{Feature masked autoencoder (fMAE).} Similar to MAE~\cite{mae}, the model is forced to learn better representations by partially masking out the input. Different from MAE, fMAE randomly masks out entire rows before encoding and fills in the condition at the masked positions; see Fig.~\ref{fig:fmae}. Specifically, the rows in the feature map are randomly masked out by $x\%\in \mathcal{N}_{truncated}(max=1, min=0.5, mean=0.75)$. In the main model training, the condition is the projected image feature.

\input{figures/fMAE}
\textbf{Image decoder.} An image decoder consisting of four deconvolutional layers is used to generation/reconstruct the images. 
Unlike Vec2Face, we observe that 
the patch-based discriminator used in Vec2Face is dropped from Vec2Face+ because its computational cost was too high for its incremental quality improvement.

\textbf{Loss function.} The training objective function includes an image reconstruction loss, an identity loss, and a perceptual loss. The reconstruction loss is:
\begin{equation}
    \mathcal{L}_{rec} = ||\mathcal{IM}_{rec}, \mathcal{IM}_{gt}||^2_2,
\end{equation}
which calculates the pixel-level difference between the reconstructed image $\mathcal{IM}_{rec}$ and the ground truth image $\mathcal{IM}_{gt}$. The identity loss is measured by cosine distance:
\begin{equation}
    \mathcal{L}_{id} = 1 - \frac{f_{rec} \cdot f_{gt}}{||f_{rec}||\ ||f_{gt}||},
\end{equation}
where the $f$ is the feature vector extracted by a pre-trained FR model, meaning that the reconstructed image should be close to the ground truth image in the face feature space. Lastly, we use the perceptual loss~\cite{lpips} with the VGG~\cite{vgg} backbone, which can be written as:
\begin{equation}
    \mathcal{L}_{lpips} = \sum_l \| w_l \odot (f_l(\mathcal{IM}_{rec}) - f_l(\mathcal{IM}_{gt})) \|_2^2
\end{equation}
where $f_l(\cdot)$ denotes the feature map extracted from layer $l$ of a pretrained VGG and $w_l$ represents the learned channel-wise importance applied to the squared difference of the features at layer $l$. This loss ensures the correct face structure at the early training stage. The total loss is:
\begin{equation}
    \mathcal{L}_{total} = \mathcal{L}_{rec} + \mathcal{L}_{id} + \lambda\mathcal{L}_{lpips}
\end{equation}
The default $\lambda$ is 0.2.

The pose control model has a frozen main model, a 4-layer CNN, and LoRA parameters. The pose control model is trained using the same objective functions as the main model, but the condition in the fMAE is the feature of an image with five face landmark points extracted by the 4-layer CNN block.

\subsection{Controllable image generation}
\label{sec:control}
Facial attribute control is key to increasing intra-class variation. Vec2Face~\cite{vec2face25} used a gradient-based algorithm, attribute operation (AttrOP), to achieve quality and pose control. 
This approach is not computationally efficient 
when generating extreme attributes; \emph{e.g.}, profile pose.
Therefore, we introduce an explicit pose control approach to mitigate this weakness. Moreover, we observe that generating images with a large head pose by using AttrOP is the main cause of intra-class identity inconsistency in the HSFace10K, so this pose control algorithm also improves intra-class identity consistency.

\input{algo/attrop}
\textbf{AttrOP.} This is a post-training generation algorithm. It combines the gradient descent algorithm and attribute evaluators to iteratively update the sampled vector until the generated images fit the target attribute conditions, as shown in Algorithm~\ref{algo:featop}. First, a target image quality $Q$, a target pose angle $P$, a pretrained pose evaluation model $M_{pose}$~\cite{pose-model}, a pretrained quality evaluation model $M_{quality}$~\cite{magface}, and an FR model $M_{FR}$~\cite{arcface} need to be set and prepared. Then, given an identity vector $v_{id}$ and a perturbed identity vector $v_{im}$, Vec2Face+ generates a face image from an adjusted vector $v'_{im}$ and computes the following loss functions:
\begin{equation}
\begin{aligned}
\mathcal{L}_{attrop}&=\mathcal{L}_{id}+\mathcal{L}_{quality}+\mathcal{L}_{pose} \text{ where}, \\
    \mathcal{L}_{id}&=1-CosSim(M_{FR}(IM), v_{id}), \\
    \mathcal{L}_{quality}&=Q-M_{quality}(IM), \\
    \mathcal{L}_{pose}&=abs(P-abs(M_{pose}(IM))),
\end{aligned}
\label{eq:attrop}
\end{equation}
Because both $M_{pose}$ and $M_{quality}$ are differentiable, gradient descent can be used to adjust $v'_{im}$ to minimize $\mathcal{L}_{attrop}$. Finally, the adjusted $v'_{im}$ is used to generate images that exhibit the desired pose and image quality.

\input{figures/pose_control}

\textbf{LoRA pose control.} Despite the usefulness of AttrOP, its iterative manner can be time-consuming and sometimes compromise the identity (see the second row of Fig.~\ref{fig:dataset}) when the target attribute is rarely represented in the training set. For example, it takes more than 20 hours to generate 200K images on an NVIDIA L40S. Inspired by~\cite{instantid}, we use face landmark points to guide the image generation to achieve a one-time generation. 
Specifically, we use a 4-layer CNN to extract the feature of the landmark images to serve as a condition in fMAE and apply LoRA~\cite{lora} to adapt the model weights to achieve control of head pose, see Fig.~\ref{fig:pose_cond}. Fig.~\ref{fig:pose_control} indicates the efficacy of the proposed method in pose control and identity preservation. Importantly, it takes less than 30 minutes to generate 200K images on an NVIDIA L40S, which is over 40x faster than using AttrOP.

In dataset assembly, we use both approaches as we notice that AttrOP effectively increases the attribute variation in general, and pose control enabled by LoRA efficiently generates images with diverse head poses.

\input{figures/pose_control_example}

\subsection{Inference: Vector sampling and Image generation}
A well-trained Vec2Face+ converts vectors to images while preserving the relationship of the vectors. This allows us to control the separability of the generated images by applying control in vector space. Specifically, generating a dissimilar image pair can be achieved by sampling two vectors with a low cosine similarity. An opposite approach can generate a similar image pair. Hence, the dataset generation consists of an identity vector sampling and an image vector generation.

\textbf{Identity vector sampling.}
Vec2Face uses a learned PCA to sample vectors and the AttrOP algorithm to promise image quality in the initial image generation, which is redundant because the final image quality is totally controlled by the AttrOP algorithm.
To reduce this redundancy, Vec2Face+ directly samples vectors from a Gaussian distribution and then applies AttrOP to control both image quality and identity. This ensures that the generated image is of good quality while its extracted feature is close to the sampled vector. Interestingly, we notice that, in the high-dimensional (\emph{i.e.}, 512) vector space, most of the randomly sampled vectors naturally fit the similarity condition due to the sparsity. Practically, we can easily sample 4M identity vectors that have a similarity lower than 0.3 to each other.

\textbf{Image vector generation.}
Once we have the initial identity images, a simple perturbation strategy is applied for intra-class image generation. First, a FR model is used to extract the features of identity images. Then, a set of random noise vectors are sampled from Gaussian distributions with $\mu=0$ and $\sigma=\{0.3, 0.5, 0.7\}$. Adding these noise vectors to identity features forms the perturbed vectors, where the cosine similarity between the perturbed vectors and their identity features is at least 0.5. Lastly, the vector norm is controlled between 18 and 24 to avoid corrupted face images.

\subsection{VFaces generation}
\label{sec:vfaces}
Utilizing the aforementioned characteristics of Vec2Face+, we generate VFace datasets, a set of datasets consisting of images from synthetic identities. The dataset construction has three image bases generated from 1) randomly sampled vectors and perturbations, 2) AttrOP with mixed pose conditions, and 3) LoRA pose control with profile face landmarks.

\textbf{Image base generated from random sampling.} There are 300K identity vectors $\mathcal{S}$ sampled from a Gaussian distribution, $\mathcal{S}=\{v_{id}\sim\mathcal{N}(0, 1)\,|\, v_{id}\in\Re^{512} \}$. After this, the identity images $\mathcal{I}$ are generated as follows.
\begin{equation}
    \mathcal{I}=\{AttrOP(\mathcal{G}(v_{id}))\,|\, Sim(v_{id}, f_{\mathcal{I}}) > 0.9\,\wedge\,Q(\mathcal{I}) > 26\}
\end{equation}
where $\mathcal{G}$ is the Vec2Face+ model, $f_{\mathcal{I}}$ is the feature of identity images extracted by an FR model, and $Q_{\mathcal{I}}$ is the quality value measured by MagFace~\cite{magface}. This promises a good image quality and identity preservation. The image generation for each identity has similar but looser constraints. For each identity, we generate 50 perturbed vectors, where 40\% are from $\mathcal{N}(0, 0.3)$, another 40\% are from $\mathcal{N}(0, 0.5)$, and the rest are from $\mathcal{N}(0, 0.7)$. After normalizing the perturbed vectors, $v_{p}$, the image generation is described below.
\begin{equation}
    \mathcal{P}=\{AttrOP(\mathcal{G}(v_{p}))\,|\, Sim(f_{\mathcal{I}}, f_{\mathcal{P}}) > 0.7\,\wedge\,Q(\mathcal{P}) > 24\}
\end{equation}
This constraint allows more diverse images within each identity folder while not falling too far from the original identity in feature space. In total, 15M images are generated in this process, but the images are mostly frontal with a small variation in facial attributes, as shown in the first row of Fig.~\ref{fig:dataset}. Hence, we use AttrOP to increase the intra-class attribute variation.

\input{figures/dataset_example}
\textbf{Image base generated from AttrOP with mixed pose conditions.} 
Combined with the attribute estimators, the gradient-based vector searching algorithm, AttrOP, can obtain vectors that can be used to generate images with the designed facial attributes. Due to the uncertainty of the gradient descent algorithm, we observe that simply changing the target pose angle can increase the variation of other attributes as well. Examples are at the second row of Fig.~\ref{fig:dataset}. Based on this observation, we use $yaw=\{30^\circ, 40^\circ, 50^\circ, 60^\circ, 70^\circ, 80^\circ\}$ as the condition and the pose estimator is SixDRepNet~\cite{pose-model}. The other conditions are the same as those used in random sampling. We run AttrOP for at most 30 iterations because of two reasons: 1) The main goal is to increase the variation of other attributes by pose-related vector searching, so obtaining the specific pose angle is not mandatory; 2) Searching vectors that can obtain images with extreme head pose (\emph{i.e.}, $> 60^\circ$) is time-consuming, so an early stopping strategy  increases efficiency. Eventually, we use this approach to generate 6M images with 20 images per identity.

\textbf{Image base generated from LoRA pose control.} Applying LoRA pose control introduced in Section~\ref{sec:control} makes the generation of images with profile pose efficient. We randomly select 30 images per identity from the image base generated by random sampling as the candidates. Then, the features of these candidates and two images with profile pose landmarks are used to generate face images with a profile pose. This approach is more identity-preserved than AttrOP when generating face images with a large pose angle, as shown in Fig.~\ref{fig:dataset}.

\textbf{VFace datasets and identity leakage check.} The proposed VFace datasets have four scales: 10K, 20K, 100K, and 300K. To be consistent with the previous work, we randomly replaced 40 images in the first image base with the generated 40 images in the second and third image bases, so that each identity has 50 images. VFace10K and VFace20K have the images of the first 10K and 20K identities. As for VFace100K and VFace300K, we apply DBSCAN\footnote{https://scikit-learn.org/stable/modules/generated/sklearn.cluster.DBSCAN.html} to reduce the dataset scale to 4M and 12M for a fair comparison with WebFace4M~\cite{webface260m}. Specifically, we use a pre-trained FR model to extract image features and use DBSCAN with the cosine distance metric to drop the outliers in each identity folder. The other settings of DBSCAN follow the default values. We check the identity leakage by leveraging a pre-trained FR model. First, we calculate the identity features of WebFace4M as it is the training set. Second, we calculate the similarity between the features of generated images and real identities. Last, generated images are dropped if any of the corresponding pairs have a similarity value higher than 0.4.
This process ensures there is no identity leakage from the training set for Vec2Face+ into the images generated by Vec2Face+. 
\section{Experiments}
\label{sec:exp}

\subsection{Experiment details}
\textbf{Vec2Face+ training set.} We train Vec2Face+ with images from WebFace4M~\cite{webface260m}, whose 4M images from 200K identities provide sufficient training data. Note that Vec2Face+ training does not require identity labels, so in principle any face dataset can be used.

\textbf{Feature extractor.}  ArcFace-R100~\cite{arcface} trained with Glint360K~\cite{glint360k} is used to extract  face features. The training configurations are those used in the insightface repository\footnote{https://github.com/deepinsight/insightface/}. Unless otherwise specified, it is referred to as ``pre-trained FR model''  in this paper.

\textbf{Test sets.}
1) \textit{five standard test sets}: LFW~\cite{lfw}, CFP-FP~\cite{cfpfp}, AgeDB-30~\cite{agedb-30}, CALFW~\cite{calfw} and CPLFW~\cite{cplfw}. LFW has 6,000 genuine and impostor pairs, testing general FR performance. CFP-FP is a subset of CFP, comprising 7,000 pairs with large variations in yaw angle. Similarly, CPLFW encompasses 6,000 image pairs with large variations in both yaw and pitch angles. AgeDB-30 is the subset of AgeDB with a 30-year age gap in each pair, challenging FR models in age variation. CALFW also emphasizes the challenge of  large age gap. 2) \textit{Two attribute-oriented test sets}: Hadrian~\cite{hadrian-eclipse} and Eclipse~\cite{hadrian-eclipse}. Hadrian and Eclipse are assembled from MORPH~\cite{morph} with 6,000 genuine and impostor pairs. The images are controlled with frontal image pose, neutral facial expression, and generally good image quality. 
These two test sets challenge FR models on facial attribute variations not emphasized in other test sets. 3) \textit{Two large-scale test sets:} IJBB~\cite{ijbc} and IJBC~\cite{ijbc}. Both IJBB and IJBC have over 100K still images and over 5,000 videos addressing recognition tasks. This paper only considers still images in the verification task. 4) \textit{Two similar-looking persons test sets}: Twins-IND~\cite{hadrian-eclipse} and DoppelVer-ViSE~\cite{Doppelver}. Twins-IND has 6,000 image pairs from Twins Challenge Dataset~\cite{nd-twins}, focusing on identical twins.
DoppelVer-ViSE contains 35,000 web-scraped image pairs, focusing on general ``doppelgangers.'' Note that, instead of using the original images in DoppelVer-ViSE, we apply img2pose~\cite{img2pose} to crop and align the images to 112$\times$112 to ensure consistency of the performance evaluation. 5) \textit{Two bias-aware test sets}: BA-test~\cite{ba-test} and BFW~\cite{bfw}. The descriptions of these two test sets are in Section~\ref{sec:bias}.

\input{table/performance}
\textbf{Experiment settings.} The FR models trained with VFace datasets use the ArcFace~\cite{arcface} loss and SE-IResNet50~\cite{resnet, SE-attention} backbone. The images are resized to 112$\times$112 with horizontal random flip, random crop, low resolution, random erase, and photometric augmentations applied. As for training settings, the FR models are trained for 40 epochs with the SGD optimizer. 
The learning rate starts at 0.1 and decays at 18, 28, and 35 epochs. The models of the reproduced results in Table~\ref{tab:performance} and Table~\ref{tab:sl-acc} are trained using the same settings.

\input{table/similar-looking}
\textbf{Evaluation protocol of test sets.} Besides IJBB, IJBC, BFW, and BA-test, the aforementioned test sets use the same evaluation protocol suggested in~\cite{lfw}. Following~\cite{arcface}, we use true positive rate (TPR) at false positive rate (FPR) equal to 1e-4 (\emph{i.e.}, TPR@FPR=1e-4) to measure the model performance on IJBB and IJBC. As for BFW and BA-test, we report TPR@FPR=1e-1 for each model.

\subsection{Model accuracy evaluation}
\textbf{Comparisons on five standard test sets.} Table~\ref{tab:performance} shows that, of synthetic training sets with less than 1M images, VFace10K results in a trained FR model with the highest accuracy across all test sets,
\emph{i.e.}, 99.35\% on LFW, 93.56\% on CFP-FP, 88.03\% on CPLFW, 94.33\% on AgeDB-30, and 94.17\% on CALFW. On average, the accuracy is 1.59\% higher than the second-best (92.30\%), showcasing the efficacy of the proposed method. With the same size as CASIA-WebFace, VFace10K achieves comparable accuracy on LFW and AgeDB-30, and better accuracy on CALFW.
Moreover,  scaling VFace up to 4M and 12M results in an FR model that achieves a higher accuracy than one trained with CASIA-WebFace. This is the first synthetic training set to enable higher accuracy than CASIA-WebFace on the five standard test sets.

\textbf{Comparisons on Hadrian, Eclipse, and large-scale test sets.} We reproduce the accuracy on Hadrian, Eclipse, IJBB, and IJBC by retraining the FR models with the available datasets. With 10K identities, the proposed VFace10K achieves on par or better accuracy than SOTA methods. Compared to CASIA-WebFace, the FR model obtains higher accuracy on IJBB and IJBC, but a large accuracy gap (7.17\% and 2.89\%) exists on Hadrian and Eclipse. After scaling the size to 12M, VFace300K achieves a higher accuracy on Eclipse than CASIA-WebFace, but there is still a gap on Hadrian.

\input{table/bias}
\textbf{Comparisons on similar-looking persons.} Identity is typically defined by cosine similarity~\cite{dcface, idiff23, arc2face, vec2face25}, where each identity has a low similarity to all other identities. This definition raises a concern for the model's performance on tasks involving similar-looking persons. In Table~\ref{tab:sl-acc}, VFace10K achieves better accuracy than the SOTA methods on DoppelVer-ViSE, and scaling further increases the accuracy. Despite the good performance on doppelganger pairs, only 1 out of 11 synthetic datasets enables a matcher that achieves accuracy better than random guessing (\emph{i.e.}, 50\%) for identical twins, whereas matchers trained on real datasets do not have this weakness. This reveals a severe problem with the approach currently used to define identity for synthetic datasets. 
Real face image datasets have persons with facial similarities due to sibling relationship, parent relationship, and even monozygotic twin relationship. Such concepts do not exist in generation of current synthetic datasets.  We speculate that this is the cause of the poor performance of synthetic training sets on similar-looking-persons test sets. Hence, more research is needed on the definition of identity used in synthetic training sets.

\input{table/ablation}
\textbf{Ablation study of the dataset assembly process.} We quantitatively analyze the effect of each process described in Section~\ref{sec:vfaces} by reporting the model accuracy on LFW, CFP-FP, CPLFW, AgeDB-30, and CALFW. The observations from the results in Table~\ref{tab:ablation} are as follows.

Generating a dataset via random sampling can achieve competitive accuracy on LFW, but not on pose and age. This is mainly due to the lack of variations in facial attributes. AttrOP enables the control of facial attributes, which increases not only the variation in pose but also in other attributes. As a result, it increases the average accuracy by 2.62\%. Different from AttrOP, the LoRA pose control method specifically increases the variation in pose, resulting in a higher accuracy on CFP-FP. The average accuracy has a 2.55\% increase. Lastly, combining these two, the proposed dataset achieves the highest accuracy on all test sets.

\subsection{Model bias evaluation}
\label{sec:bias}
Accuracy disparity across demographic groups is a serious  issue~\cite{gender-bias, race-bias, Hoggins2019, Lohr2018, Santow2020, Vincent2019}, but has not been analyzed in synthetic datasets. To address this, we compare the model's performance on two test sets, BFW~\cite{bfw} and BA-test~\cite{ba-test}. BFW balances the number of identities and images for each demographic group, comprising 20K images from 800 identities across 8 demographic groups. BA-test argues that, in face verification, balancing important attributes in the dataset is more important than image and identity numbers, so it balances head pose, image quality, and face brightness. We use its benchmark test set, which further balances the number of identities and images, encompassing 3,600 images from 8 demographic groups. These two test sets provide bias-aware evaluation in different aspects, and the demographic groups follow the original dataset settings.

We report the TPR for each demographic group, where the threshold is calculated with an FPR equal to 0.1 based on the similarity value of all impostor pairs. 
Since it is hard to obtain insights from low-accuracy results, we ignore those TPR values below 60\%.
The analysis of Table~\ref{tab:demog-acc} consists of three parts: 1) gender accuracy disparity, 2) the group with the highest accuracy, and 3) the group with the lowest accuracy. \textit{Gender accuracy disparity:} It is consistent that Indian Male (IM) has lower accuracy than Indian Female (IF) and Black Male (BM) has higher accuracy than Black Female (BF) in these two test sets. However, the pattern of Asian and White is opposite in the two test sets, showcasing that controlling for different factors in test set results in different patterns. \textit{Group with the highest accuracy:} In both test sets, the Indian group has the highest accuracy. \textit{Group with the lowest accuracy:} The lowest accuracy mainly occurs in the Asian group. 

We conclude the observations mainly based on the results of BA-test dataset, as the facial attributes are more tightly controlled in it. One, the model trained with the real dataset has less gender bias in general. Two, the difference between the highest and lowest accuracy values on synthetic datasets is significantly larger than that on the real dataset, indicating a larger racial bias of the model trained with synthetic datasets. Both observations suggest that more attention should be paid to bias mitigation in synthetic dataset generation.

\section{Further analysis}
\label{sec:further}
Although the proposed datasets achieve state-of-the-art accuracy, there is still an accuracy gap between synthetic and real datasets, especially for twin verification. To investigate this gap, we analyze the inter-class separability and intra-class consistency for both average and edge cases. The datasets involved are the same seven datasets used in Section~\ref{sec:attr-analysis} and the proposed VFace10K (V.).

\input{figures/correlation-acc-inter}
\input{figures/correlation-acc-intra}
\subsection{Average case analysis}
We use the same metrics in Section~\ref{sec:attr-analysis} to calculate the separability degree and consistency degree. To obtain inspiration, we draw the correlation between both factors and the accuracy achieved for each dataset, as shown in Fig.~\ref{fig:corr-sep} and Fig.~\ref{fig:corr-consis}. 
The observations are summarized as follows. One, inter-class identity separability is important, but the benefit plateaus after a certain degree (\emph{i.e.}, 0.7). Two, intra-class identity consistency is a critical factor that causes the accuracy gap. Except for VFace10K, the existing synthetic datasets have an obvious gap with real datasets in identity consistency degree. Both factors have been resolved by the proposed Vec2Face+ and VFace datasets, resulting in better accuracy than other methods. However, both factors are not related to the issue of synthetic datasets yielding matchers with poor accuracy on identical twins test sets.

\subsection{Edge cases analysis}
Despite achieving a good average performance of inter-class separability and intra-class identity consistency, VFace10K still does not achieve a better performance than CASIA-WebFace. Due to this, we conduct the edge case analysis experiment by using human judgment.

\textbf{Data preparation.} Inter-class (across identity) and intra-class (within identity) groups of images were selected in
the same manner for each dataset, to be rated by human observers to assess identity
separation and identity consistency, respectively. For the inter-class analysis, the 100 most-
similar identity pairs are selected within each dataset, with any identity occurring only once in
the 100 pairs. Similarity of an identity pair is estimated by cosine similarity of the average
feature vectors for the identities, with feature vectors computed from a pretrained ArcFace~\cite{arcface} matcher. For each of the 100 most-similar identity pairs, the 4 most-similar cross-identity
image pairs are selected, with any image occurring only once in the 4 pairs. This results in 100
sets of 8 images each, to be judged by a human observer for whether the images could all
belong to one identity. For the intra-class analysis, 100 identities are selected at random for
each dataset, and for each identity, the four least-similar image pairs selected, with any image
occurring only once in the 4 pairs. This results in 100 sets of 8 images each, to be judged by a
human observer for whether the images could all belong to one identity.

\input{figures/correlation-acc-inter-human}
\input{figures/correlation-acc-intra-human}
\textbf{Experiment settings.} Participants in the experiment viewed a series of 105 slides, each containing a set of 8 face
images, and rated each set of images on a five-point Likert scale for the possibility of all 8
images in the being of the same individual. Participants viewed the slides on a desktop
computer screen in a supervised setting. The experiment followed an IRB-approved protocol,
all subjects completed an informed consent form, and participants received a \$10 gift card.

\textbf{Data processing.} Each of the 200 image sets from each dataset received at least 5 participant ratings. The mean
rating value was computed for each image set. Min-max normalization of the mean ratings was
performed across the image sets to place the means in the range of 0 to 1. Then, the average
inter-class rating and average intra-class rating were computed for each dataset. Identity
consistency of each dataset is represented by its mean intra-class rating, with 1 representing
maximum consistency. Identity separation of each dataset is represented by (1 – average inter-
class rating), with 1 representing maximum separation.

The observations from Fig.~\ref{fig:corr-sep-human} and Fig.~\ref{fig:corr-consis-human} are as follows. 1) Having similar inter-class identities in the dataset does not hurt the accuracy, which is consistent with the conclusion in the average case analysis; 2) Having intra-class identity outliers hurts the model performance; 3) Both factors are not related to the issue of synthetic datasets yielding matchers with poor accuracy on identical twins test sets. In Fig.~\ref{fig:corr-sep-human}, CASIA-WebFace has the second-lowest separability when judging the most similar inter-class identity pairs. Combining the observation in the average case analysis, it is not harmful to have similar inter-class identities as long as the average separability degree is larger than 0.7. In Fig.~\ref{fig:corr-consis-human}, excluding DigiFace, CASIA-WebFace has the best intra-class consistency. This shows that synthetic datasets have more intra-class identity outliers than CASIA-WebFace, which may cause the accuracy gap.

\section{Conclusion}
This paper explores generation of synthetic datasets for training face matching algorithms, as a means to address possible privacy concerns in using web-scraped datasets.
First, we reveal that intra-class identity consistency is an important but neglected factor in FR training set generation. To achieve a high intra-class identity consistency in the training set, we propose Vec2Face+, a model that is identity-aware, efficient for generation, and allows facial attribute control. Using this model, we generate four versions of VFace datasets, VFace10K, VFace20K, VFace100K, and VFace300K. The VFace10K achieves higher accuracy on LFW, CFP-FP, CPLFW, AgeDB, and CALFW than the previous methods of generating synthetic training sets. VFace100K and VFace300K eventually achieve higher average accuracy than a real dataset CASIA-WebFace. To our knowledge, this is the first time a synthetic dataset enables higher accuracy than CASIA-WebFace.

We further involve 8 test sets in model testing and analyze the effect of inter-class separability and intra-class identity consistency on accuracy. We observe that 1) FR models trained with synthetic identities are more biased on accuracy, 2) the similarity-based definition of identity used in current methods of synthetic dataset generation causes them to generate matchers that fail in the twin verification task, 3) identity separability is important, but its effect is saturated after 0.7, 4) existing dataset creation methods can achieve a good intra-class attribute variation but struggle in maintaining a good intra-class identity consistency. We hope these observations can help to focus research progress in this area.

\ifCLASSOPTIONcaptionsoff
  \newpage
\fi

\bibliographystyle{IEEEtran.bst}
\bibliography{main}

\section{Biography Section}
 
\vspace{11pt}

\begin{IEEEbiography}[{\includegraphics[width=1in,height=1.25in,clip,keepaspectratio]{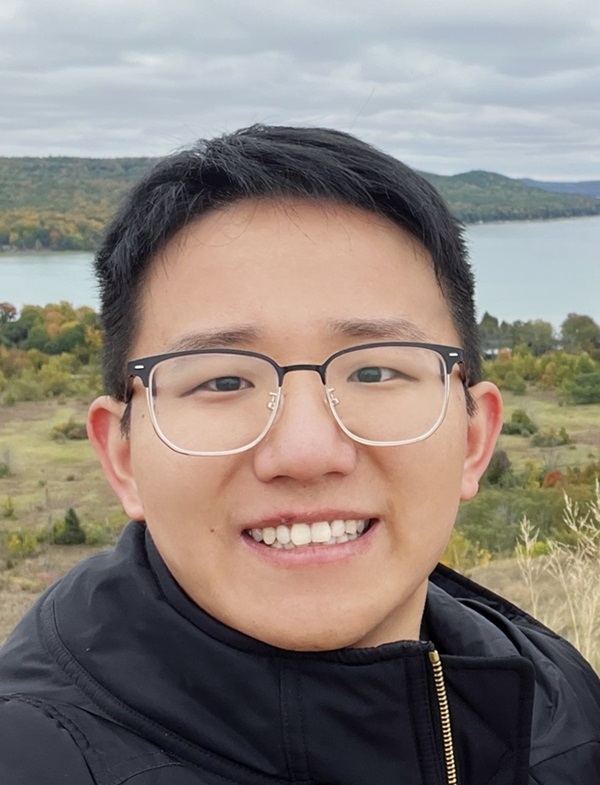}}]{Haiyu Wu} 
received the B.S. degree in electrical engineering computer engineering emphasis from Northern Arizona University in 2020 and the PhD degree in computer engineering from the University of Notre Dame. He serves as a program chair in a workshop at ICCV2025. He is a Research Scientist at Altos Labs. His research interests include computer vision and multi-modal generative models.

\end{IEEEbiography}

\begin{IEEEbiography}[{\includegraphics[width=1in,height=1.25in,clip,keepaspectratio]{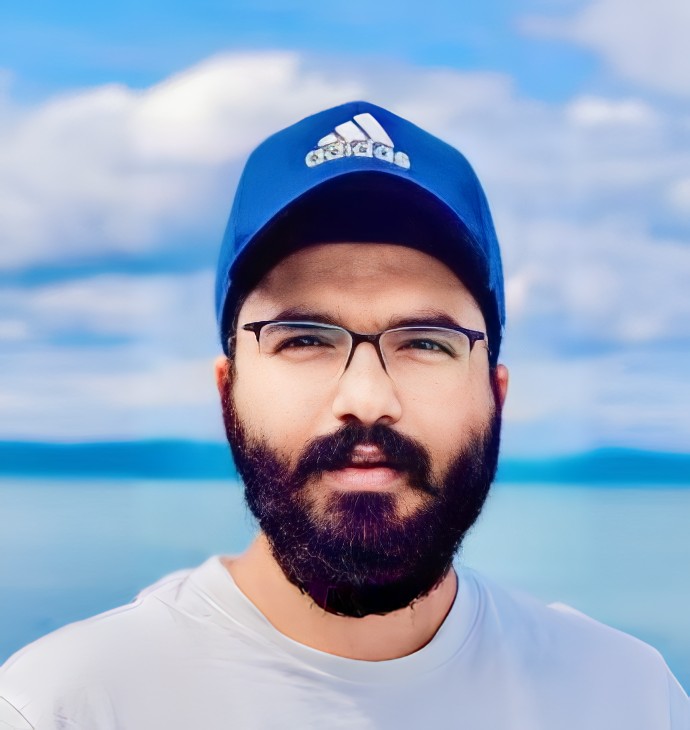}}]{Jaskirat Singh} is a Ph.D. student at the Research School of Computer Science, at Australian National University. He received his Bachelors from Indian Institute of Technology Delhi (IIT-Delhi). He also received his Masters degree in Machine learning and Computer Vision with top-honor from the Australian National University. His research interests include multi-modal fusion and reinforcement learning driven reasoning agents.

\end{IEEEbiography}

\begin{IEEEbiography}[{\includegraphics[width=1in,height=1.25in,clip,keepaspectratio]{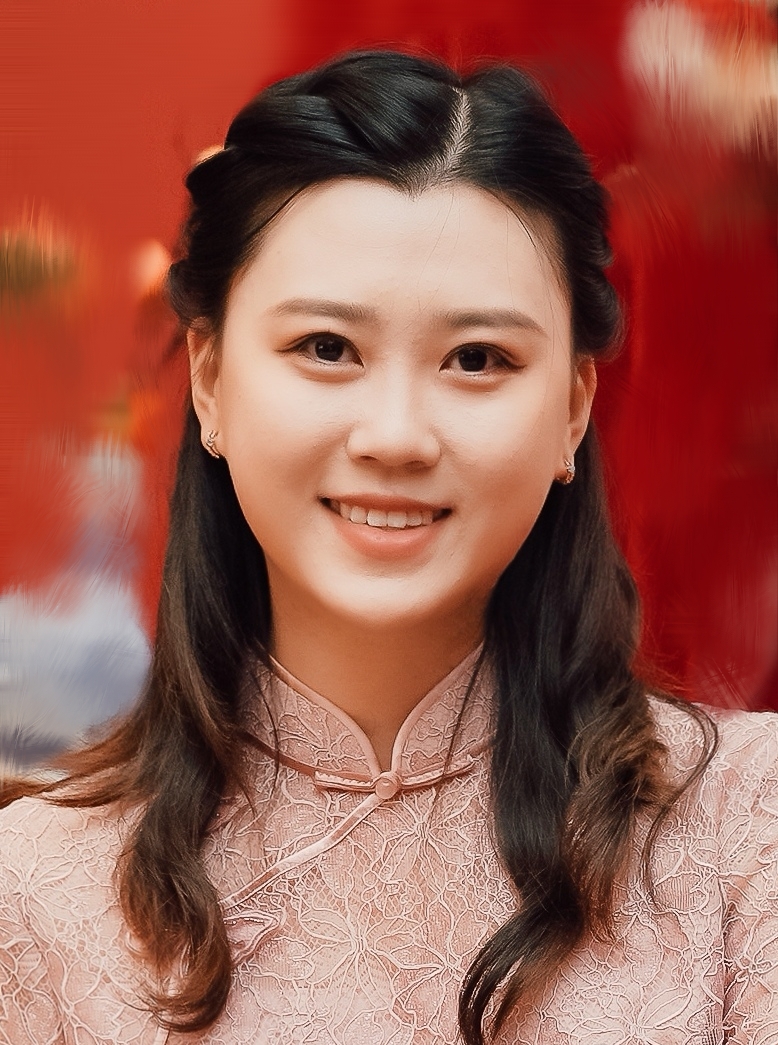}}]{Sicong Tian} received the Master's degree in mathematics and computer science from Indiana University South Bend with the award of excellence. Her research interests include improving the ethical responsibility of face recognition algorithms, computer vision and machine learning.

\end{IEEEbiography}

\begin{IEEEbiography}[{\includegraphics[width=1in,height=1.25in,clip,keepaspectratio]{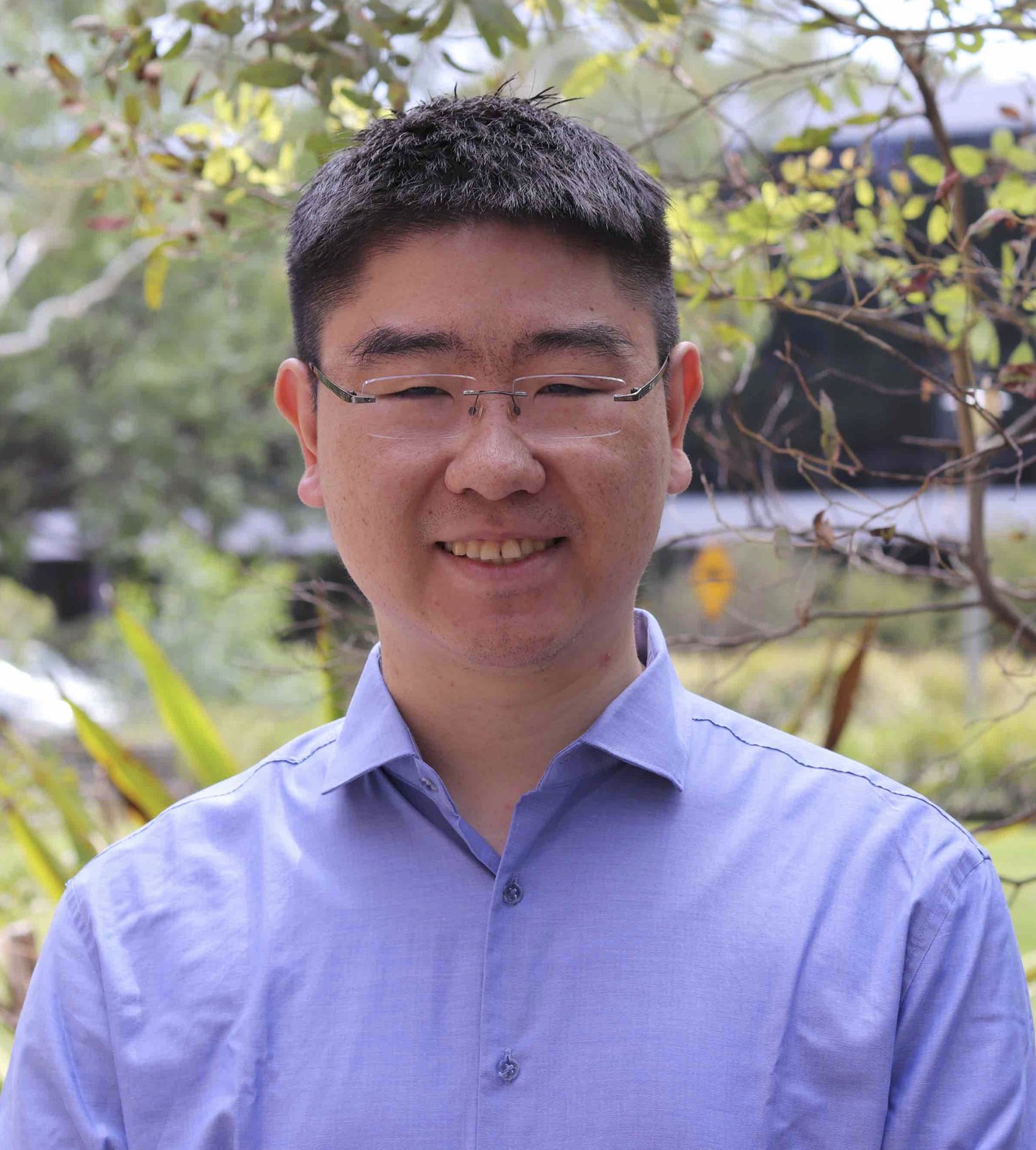}}]{Liang Zheng} (Senior Member, IEEE) received the BE degree in life science from Tsinghua University,
China, in 2010 and the PhD degree in electronic engineering from Tsinghua University, China, in 2015. He is an associate professor with the School of Computing, Australian National University. He regularly serves as Area Chair for leading conferences including ICCV, CVPR, ECCV, NeurIPS and ICML. He was Program Co-Chair for ACM MM 2024 and will be General Co-Chair for AVSS 2026. He is a Senior AE for IEEE TCSVT and an AE for ACM Computing Survey. His research interests include object reidentification, generative AI, and data-centric machine learning.

\end{IEEEbiography}

\begin{IEEEbiography}[{\includegraphics[width=1in,height=1.25in,clip,keepaspectratio]{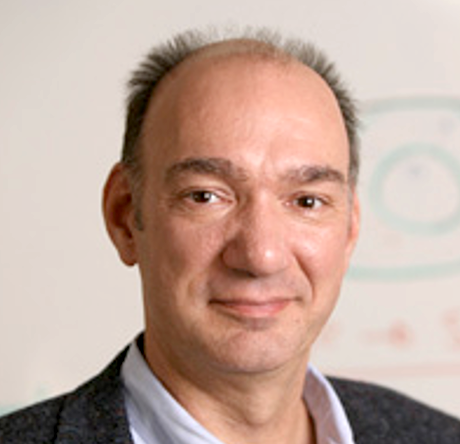}}]{Kevin W. Bowyer}
 is the Schubmehl-Prein Family Professor of Computer Science and Engineering at the University of Notre Dame.  Professor Bowyer is a Fellow of the American Academy for the Advancement of Science “for distinguished contributions to the field of computer vision and pattern recognition, biometrics, object recognition and data science”, a Fellow of the IEEE “for contributions to algorithms for recognizing objects in images”, and a Fellow of the IAPR “for contributions to computer vision, pattern recognition and biometrics”.  He received a Technical Achievement Award from the IEEE Computer Society, and both the Meritorious Service Award and the Leadership Award from the IEEE Biometrics Council.  Professor Bowyer served as Editor-In-Chief of both the IEEE Transactions on Biometrics, Behavior, and Identity Science and the IEEE Transactions on Pattern Analysis and Machine Intelligence.  
\end{IEEEbiography}

\vfill

\end{document}

%% file: figures/teaser.tex
\begin{figure}
    \centering
\includegraphics[width=\columnwidth]{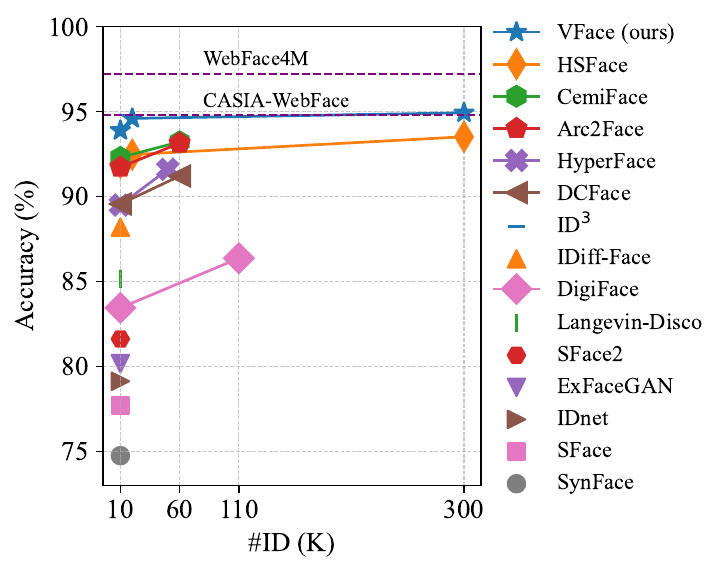}
\vspace{-8mm}
\caption{Comparing existing synthetic FR training sets of their average accuracy on five real-world test sets: LFW, CFP-FP, AgeDB-30, CALFW, and CPLFW. Generated by the Vec2Face+ method, the VFace datasets exhibit state-of-the-art and scaling performance. Notably, for the first time we report higher average accuracy than the real-world CASIA-WebFace training set.} 
\label{fig:teaser}
\end{figure}

%% file: figures/vec2face-paradigm.tex
\begin{figure}
    \centering
\includegraphics[width=\columnwidth]{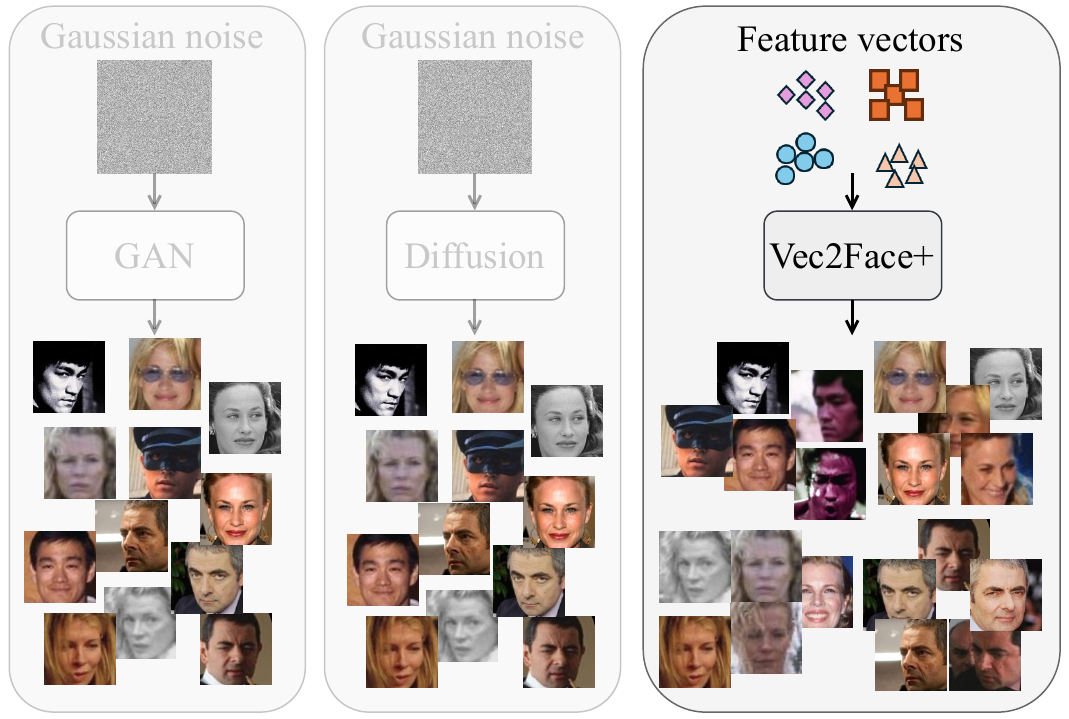}
\caption{GAN-based and diffusion-based methods combine a Gaussian noise image and conditions to generate images. This compromises the identity control in the image domain. Different from them, the proposed Vec2Face+ learns image generation by understanding vectors extracted by a feature model, such that the learned generative model can transfer the correlation of vectors into images, achieving an effective identity control.}
\label{fig:paradigm}
\end{figure}

%% file: figures/facial-attr.tex
\begin{figure*}[t]
    \centering
    \begin{minipage}{0.32\textwidth}
        \includegraphics[width=\linewidth]{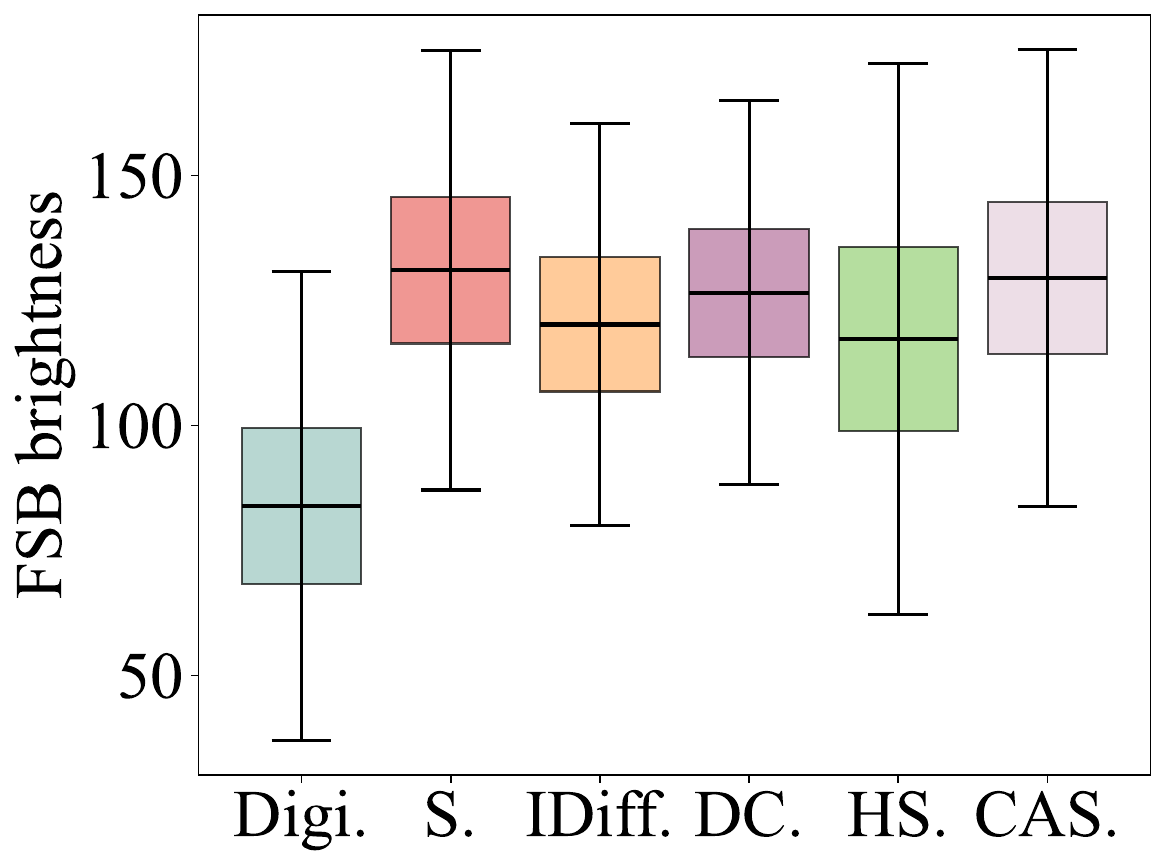}
    \end{minipage}
    \hfill    
    \begin{minipage}{0.32\textwidth}
        \includegraphics[width=\linewidth]{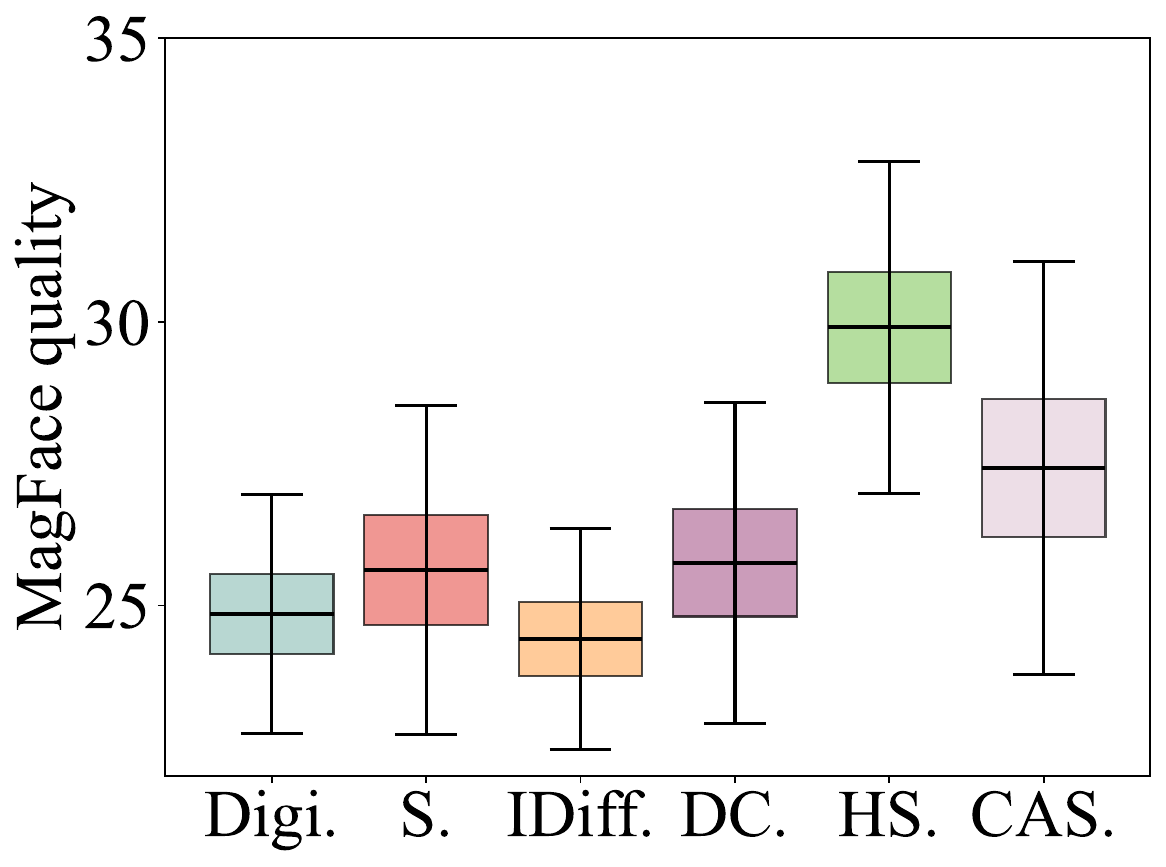}
    \end{minipage}
    \hfill
    \begin{minipage}{0.32\textwidth}
        \includegraphics[width=\linewidth]{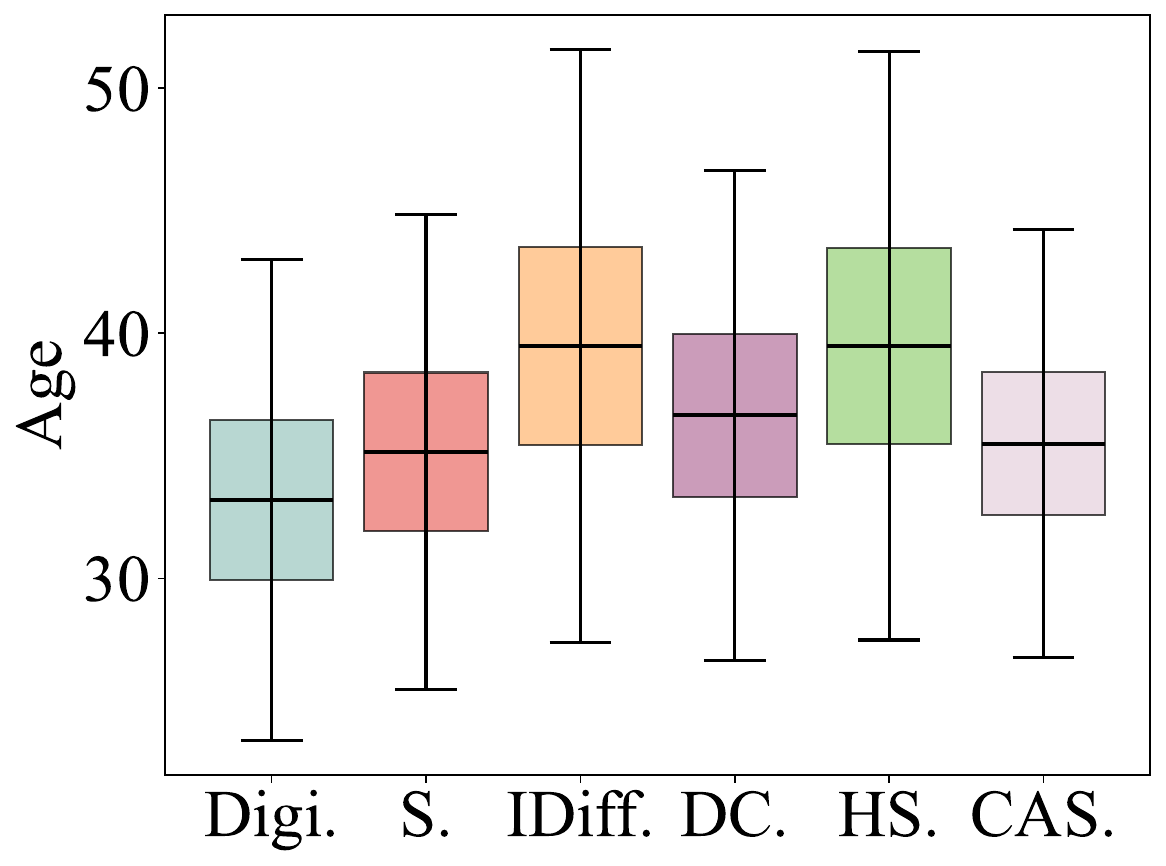}
    \end{minipage}
    \hfill
    \begin{minipage}{0.32\textwidth}
        \includegraphics[width=\linewidth]{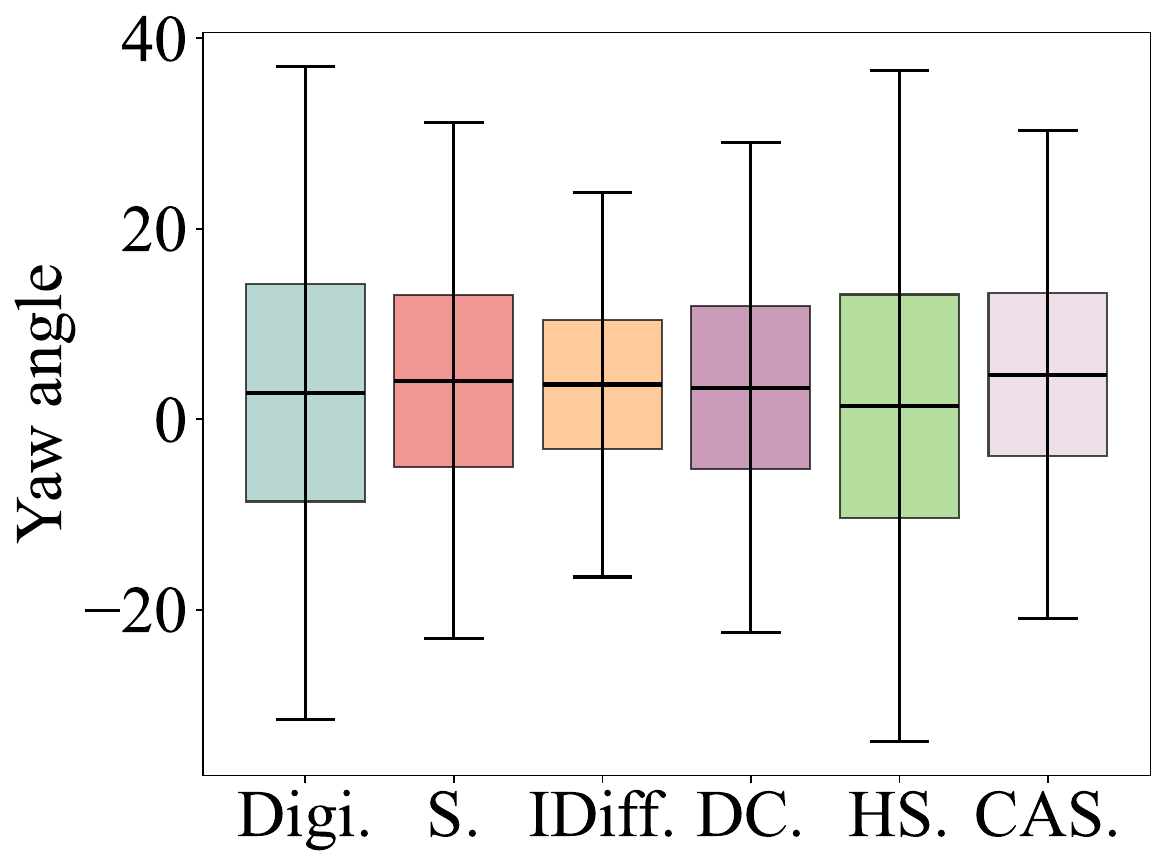}
    \end{minipage}
    \hfill
    \begin{minipage}{0.32\textwidth}
        \includegraphics[width=\linewidth]{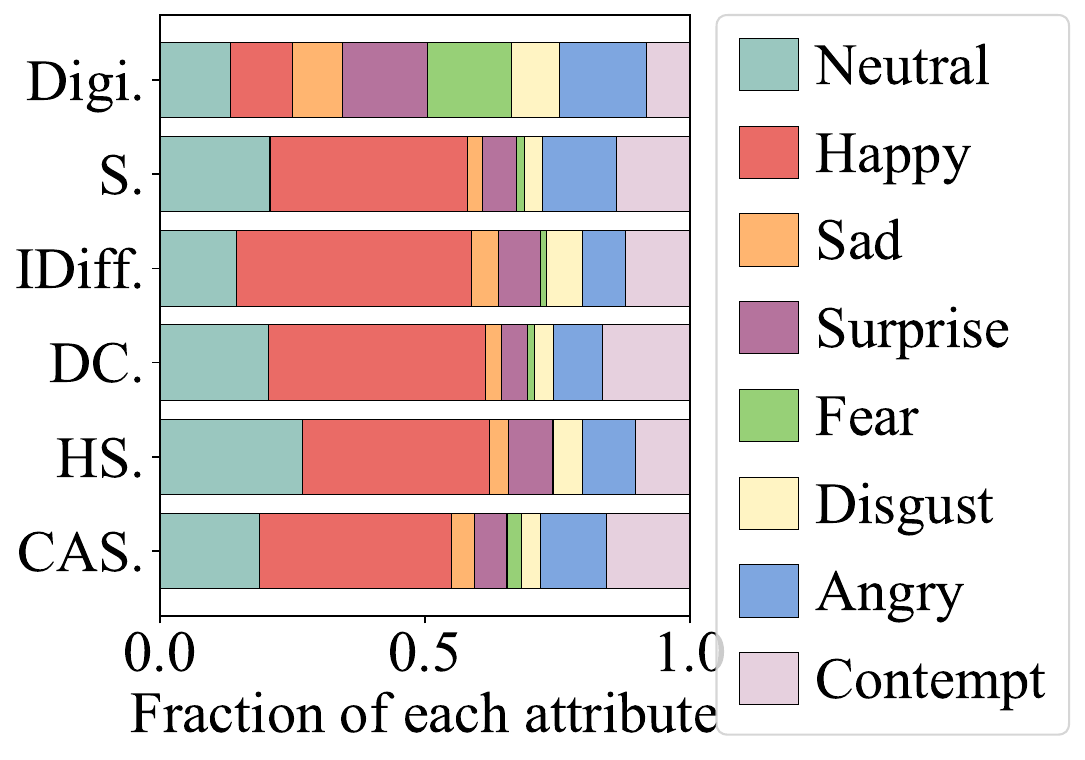}
    \end{minipage}
    \hfill   
    \begin{minipage}{0.32\textwidth}
        \includegraphics[width=\linewidth]{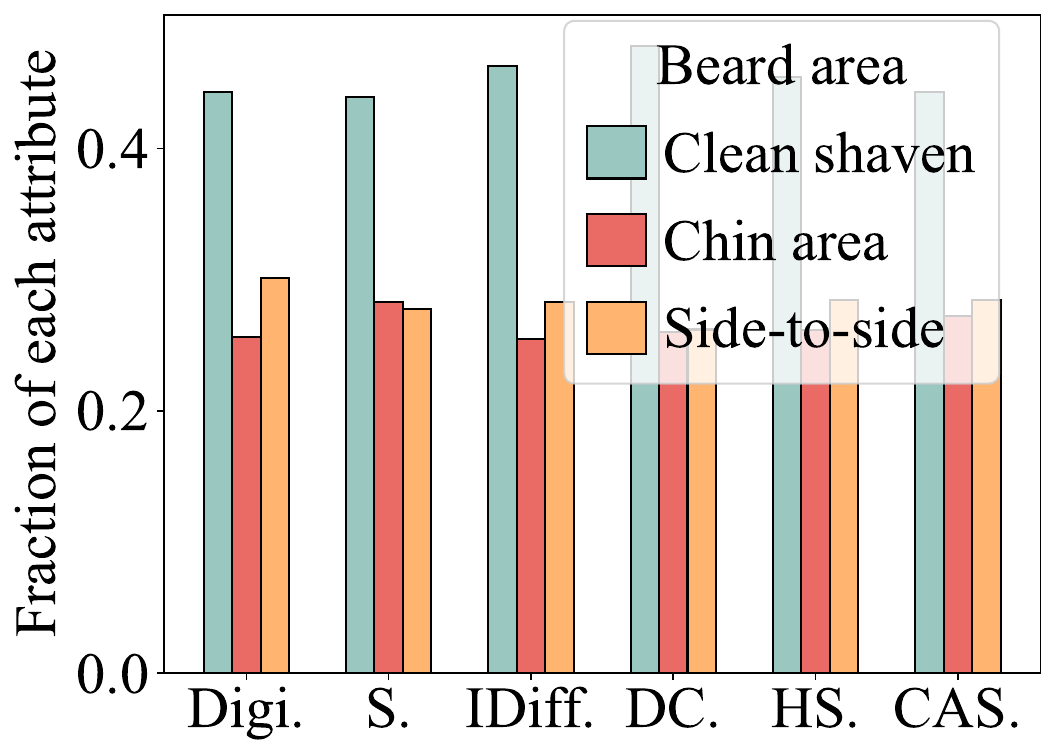}
    \end{minipage}
    \hfill
    \begin{minipage}{0.32\textwidth}
        \includegraphics[width=\linewidth]{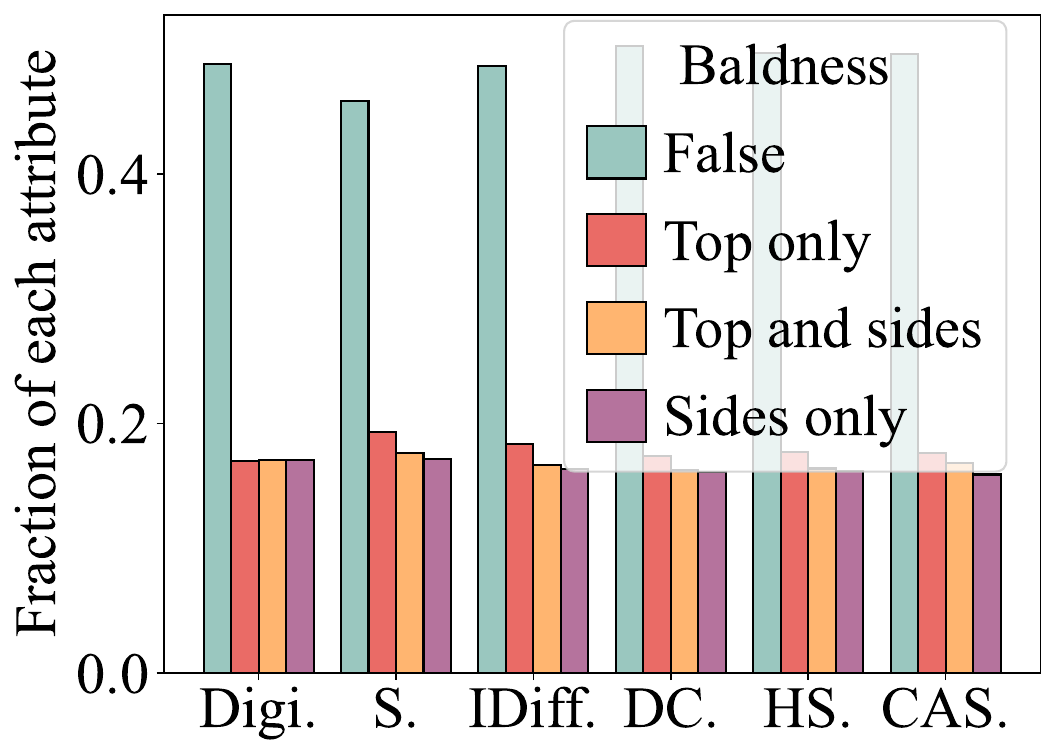}
    \end{minipage}
    \hfill
    \begin{minipage}{0.32\textwidth}
        \includegraphics[width=\linewidth]{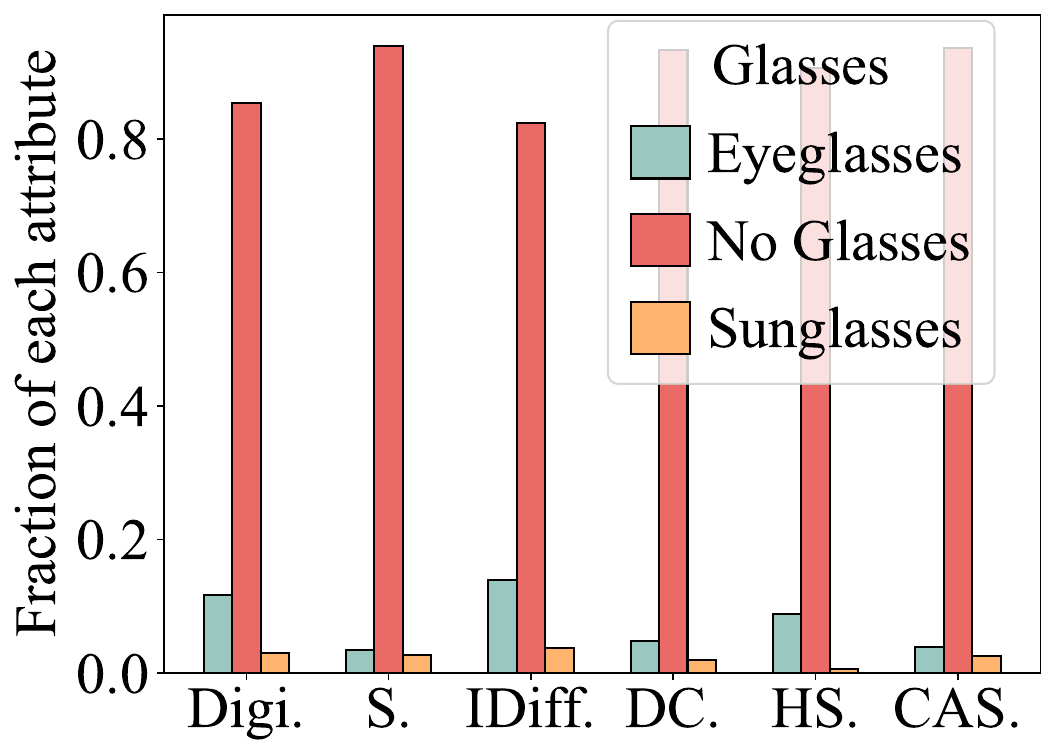}
    \end{minipage}
    \hfill
    \begin{minipage}{0.32\textwidth}
        \includegraphics[width=\linewidth]{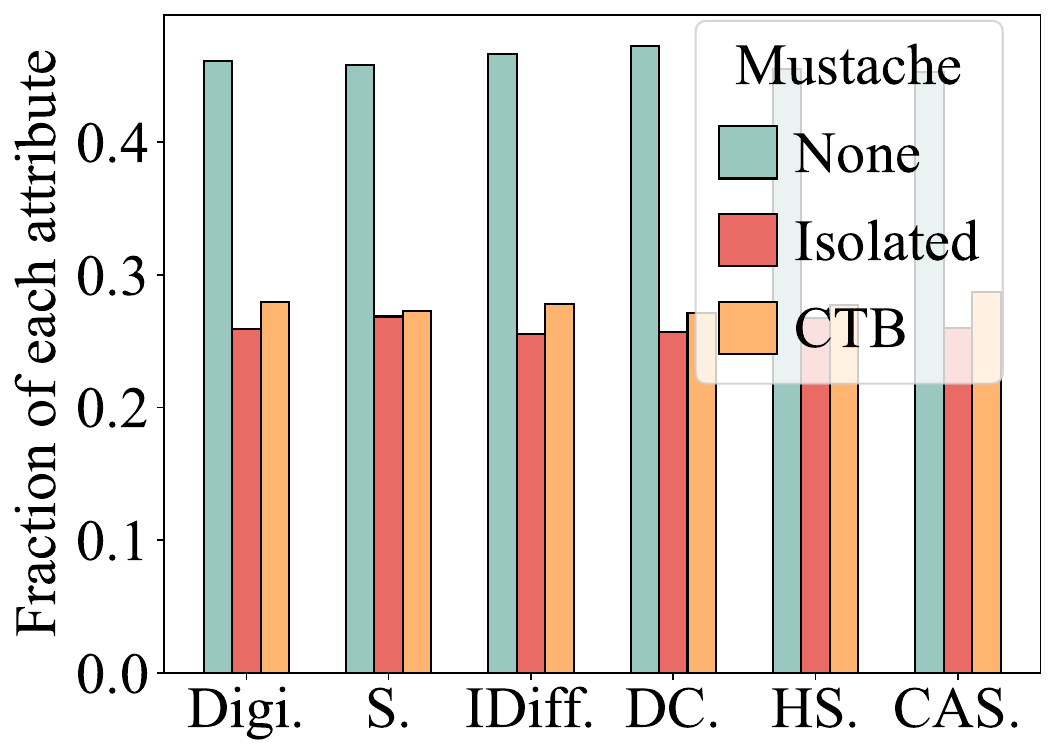}
    \end{minipage}
    \caption{Does intra-class variation decide training data effectiveness? We compare five synthetic datasets and a real dataset (CAS.), where each figure shows a comparison on a certain facial attribute. Attribute variation is measured by the whisker length or the fraction. We observe that existing methods can achieve an on-par or larger variation than the real dataset. Abbreviations: Digi. (DigiFace), S. (SFace), IDiff. (IDiff-Face), DC. (DCFace), HS. (HSFace10K), and CAS. (CASIA-WebFace). The attribute names follow the original datasets. CTB means connected to beard. 
    }
\label{fig:intra-attr}
\end{figure*}

%% file: figures/inter-sep.tex
\begin{figure}[t]
    \centering
    \includegraphics[width=\linewidth]{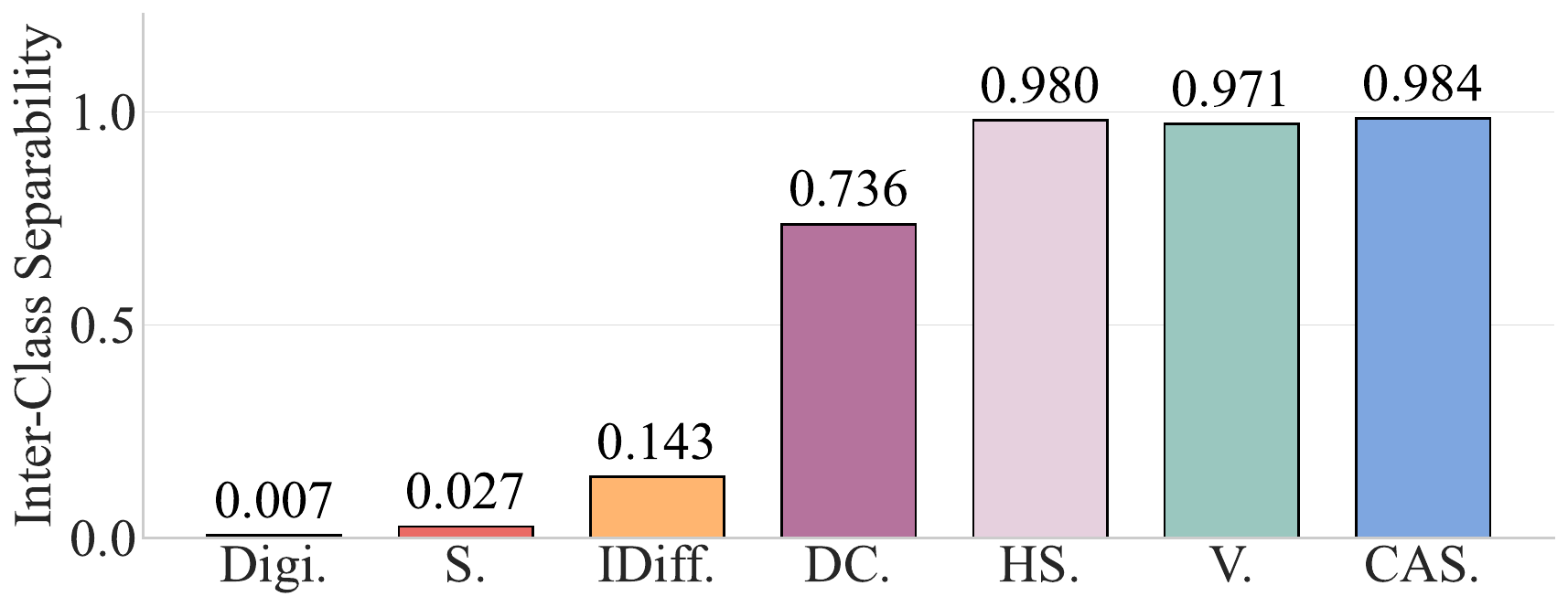}
    \caption{Comparing synthetic and real FR training sets of their inter-class separability. All datasets have 10K identities and V. is the VFace10K dataset proposed in this paper. HSFace10K and VFace10K share similar separability with CASIA-WebFace, while DigiFace has the lowest separability. Separability benefits FR accuracy (Fig. \ref{fig:corr-sep}).}
    \label{fig:inter-sep}
\end{figure}

%% file: figures/inconsistency.tex
\begin{figure}[t]
    \centering
    \includegraphics[width=\linewidth]{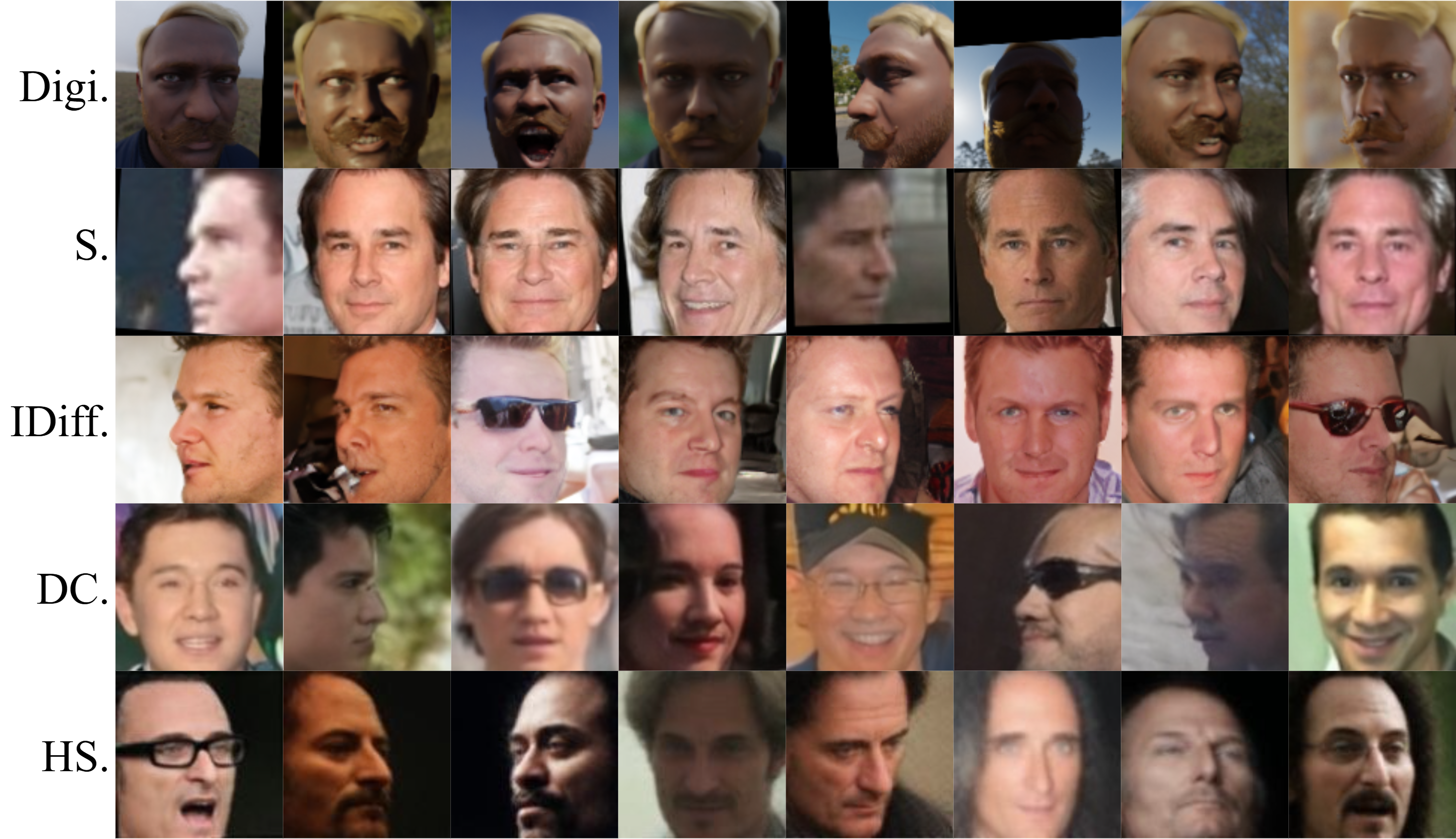}
    \caption{Examples of inconsistent identities in five synthetic datasets. Images in each row are from the same identity folder in the named dataset, where we often observe identity inconsistency. The last row shows images from VFace, showing higher consistency which benefits FR accuracy. Statistical comparisons are provided in Fig. \ref{fig:intra-consis} and Fig. \ref{fig:corr-consis}.
    }
    \label{fig:intra-inconsis}
\end{figure}

%% file: figures/intra-consis.tex
\begin{figure}[t]
    \centering
    \includegraphics[width=\linewidth]{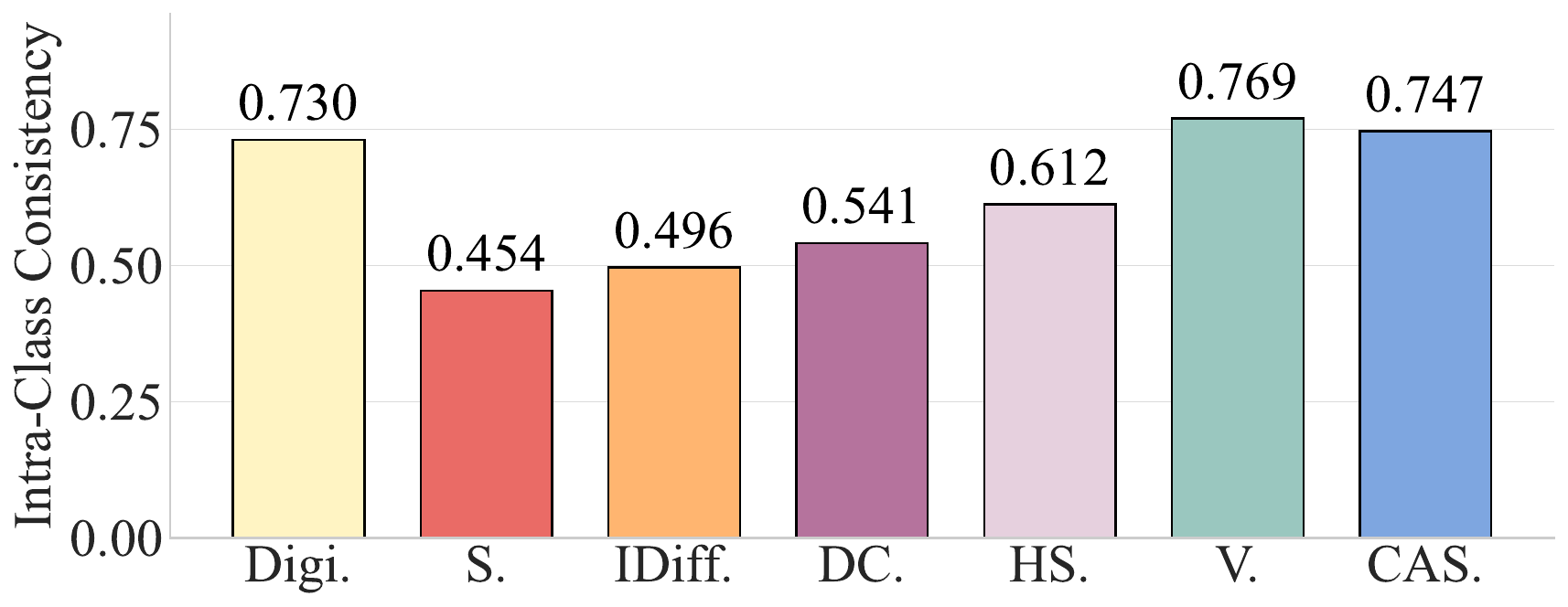}
    \caption{Comparing six datasets of their intra-class identity consistency. All datasets have 10k identities. While DigiFace has similar identity consistency with CASIA-WebFace, it has low inter-class separability according to Fig. \ref{fig:inter-sep}. Except DigiFace, all synthetic datasets exhibit identity confusion compared with CASIA-WebFace. VFace10K has improved identity consistency, which benefits FR accuracy in Fig. \ref{fig:corr-consis}. Visual examples are shown in Fig. \ref{fig:intra-inconsis}.}
    \label{fig:intra-consis}
\end{figure}

%% file: figures/vec2face-training.tex
\begin{figure}
    \centering
\includegraphics[width=\columnwidth]{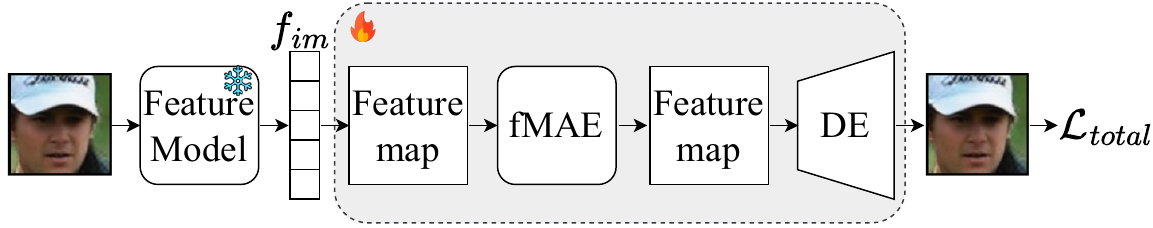}
\caption{Architecture of Vec2Face+. The input is the image feature, $f_{im}$, extracted by an FR model. After expanding the $f_{im}$ to a feature map by using two linear layers, the feature masked auto-encoder (fMAE) maps it to another feature map that is used to decode an image. The model training is supervised by the image reconstruction.}
\label{fig:vec2face-arch}
\end{figure}

%% file: figures/fMAE.tex
\begin{figure}
    \centering
\includegraphics[width=\columnwidth]{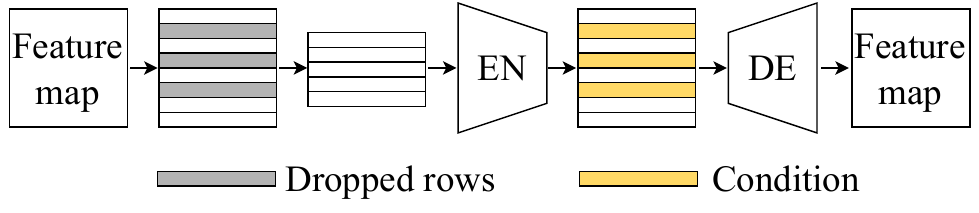}
\caption{Architecture of fMAE. The fMAE mainly adopts the idea of MAE~\cite{mae}. Different from MAE, we randomly drop rows from the feature map before encoding and filling in the condition at dropped rows' positions after encoding. As the improved Vec2Face~\cite{vec2face25}, Vec2Face+ further investigates the potential of controllability in fMAE (see Fig.~\ref{fig:pose_cond}) to improve intra-class identity consistency.}
\label{fig:fmae}
\end{figure}

%% file: algo/attrop.tex
\begin{algorithm}[t]
    \SetKwFunction{FMain}{AttrOP}
    \SetKwProg{Fn}{Function}{:}{}
    \Fn{\FMain{$v_{id}$, $v_{im}$, $M_{gen}$, $T$}}{
        \KwIn{a) $v_{id}$: sampled ID vectors, \\ \hspace{10.5mm}b) $v_{im}$: initial perturbed ID vectors, \\ \hspace{10.5mm}c) $M_{gen}$: Vec2Face+ main model, \\ \hspace{10.5mm}c) $T$: the number of iterations}
        \KwRequired{target quality $Q$, target pose $P$}
        \KwOut{a) $v'_{im}$: adjusted perturbed ID vectors}
        \DontPrintSemicolon
        Condition models: $M_{pose}$, $M_{quality}$, $M_{FR}$ \;
        Initialization $v'_{im}=v_{im}$ \;
        \For{$t=T-1,T-2, ..., 0$}{
            $IM=M_{gen}(v'_{im})$ \;
            Calculate $\mathcal{L}_{attrop}$ in Eq.~\ref{eq:attrop} \;
            $v'_{im}=v'_{im}-\lambda\nabla_{v'_{im}}\mathcal{L}_{attrop}$
        }
        return $v'_{im}$
    }
    \caption{AttrOP}
    \label{algo:featop}
\end{algorithm}

%% file: figures/pose_control.tex
\begin{figure}
\centering
\includegraphics[width=\columnwidth]{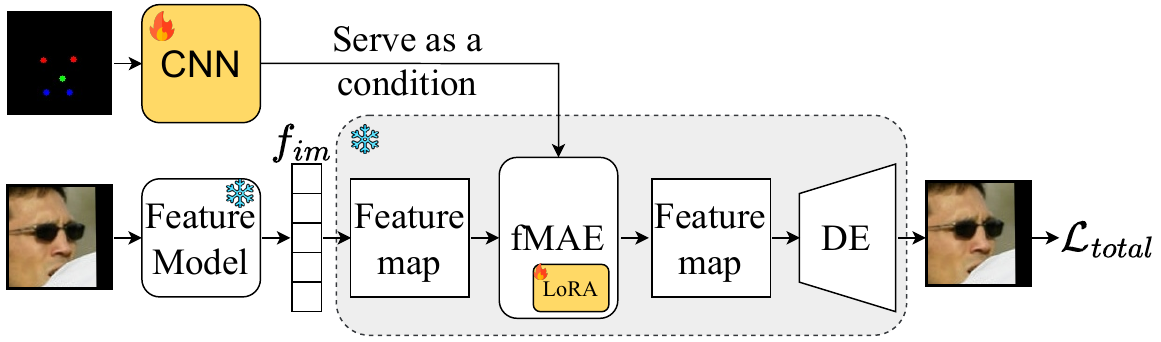}
\caption{Architecture of conditional fine-tuning. A CNN block and parameter efficient LoRA~\cite{lora} fine-tuning are used to adapt the model for explicit pose control. The CNN is used to extract the feature of a face landmark image, and this feature serves as a condition in fMAE. LoRA parameters enable the control of head pose. Examples of pose control are in Fig.~\ref{fig:pose_control}}
\label{fig:pose_cond}
\end{figure}

%% file: figures/pose_control_example.tex
\begin{figure}
    \centering
\includegraphics[width=\columnwidth]{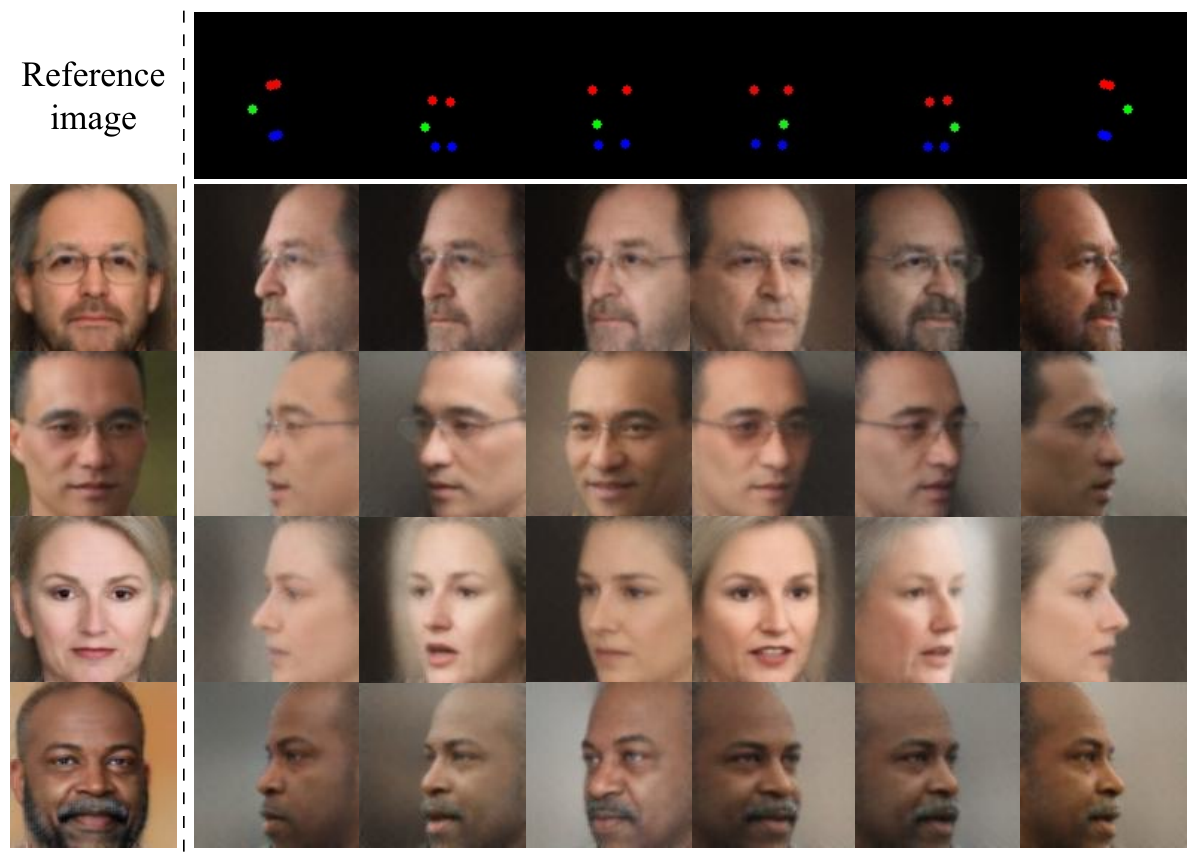}
\caption{Examples of head pose control. Leveraging the LoRA~\cite{lora} parameter-efficient fine-tuning, Vec2Face+ allows the control of the head pose by giving an image with five face landmark points. This achieves the efficient generation of face images with a profile pose.}
\label{fig:pose_control}
\end{figure}

%% file: figures/dataset_example.tex
\begin{figure}
\centering
\includegraphics[width=\columnwidth]{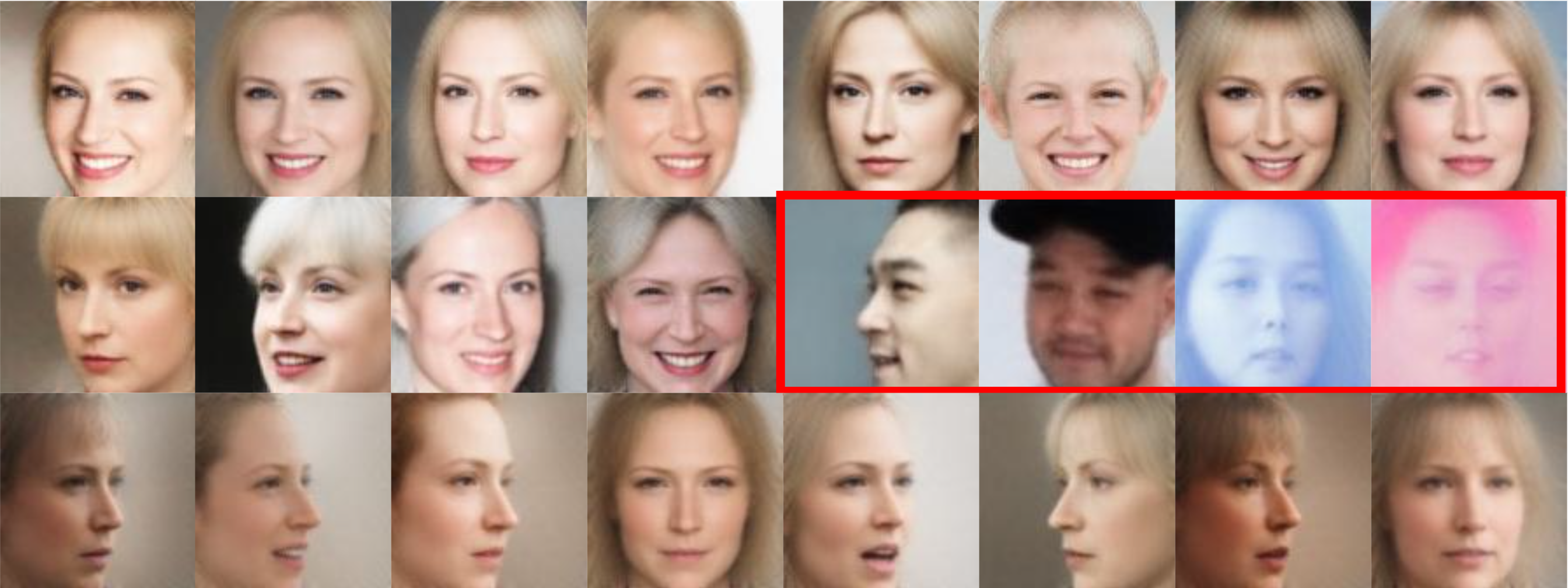}
\caption{Examples of generated images for one identity by three different strategies. Row 1: random feature perturbation. Row 2: AttrOP with mixed pose conditions, where the examples of the compromised identity are shown in the red rectangle. Row 3: images with various head poses generated by the LoRA pose control algorithm. It can generate images with large head poses while preserving the identity.}  %
\label{fig:dataset}
\end{figure}

%% file: table/performance.tex
\setlength{\tabcolsep}{1.65mm}
\begin{table*}[t]
\centering
\caption{Comparison of existing synthetic training sets on nine test sets. 
For five standard test sets, we report both average accuracy (Avg.) and difference from real datasets Casia-WebFace (CWF) and WebFace4M (W4M).
Results with $\star$ are from model trained on distributed dataset. [Keys: \textbf{best accuracy}, \underline{second best}, \better{higher than real}, \worse{lower than real}.]}
\begin{tabular}{l|c|c|c|c|c|c|l|c|c||c|c}
\hline
Training sets  & \# ims & LFW   & CFP-FP & CPLFW & AgeDB-30 & CALFW & \multicolumn{1}{c|}{Avg. (CWF / W4M)} & Hadrian$^\star$ & Eclipse$^\star$ & IJBB$^\star$ & IJBC$^\star$ \\ \hline %
SynFace \cite{synface} & 0.5M           & 91.93 & 75.03 & 70.43 & 61.63 & 74.73 & 74.75 (\worse{-20.04}/\worse{-22.48}) & - & - & - & -\\ %
SFace \cite{sface22}  & 0.6M           & 91.87 & 73.86 & 73.20 & 71.68 & 77.93 & 77.71 (\worse{-17.08}/\worse{-19.52}) & 49.47 & 49.35 & 37.90 & 41.94\\ %
DigiFace \cite{digiface-1m} & 1M           & 95.40 & 87.40 & 78.87 & 76.97 & 78.62 & 83.45 (\worse{-11.34}/\worse{-13.78}) & 47.75 & 48.78 & 5.42 & 6.10\\ %
IDiff-Face \cite{idiff23}  & 0.5M           & 98.00 & 85.47 & 80.45 & 86.43 & 90.65 & 88.20 (\worse{-6.59}/\worse{-9.03}) & 58.73 & 59.67 & 62.13 & 63.61 \\ %
ExFaceGAN \cite{exfacegan23} & 0.5M           & 93.50 & 73.84 & 71.60 & 78.92 & 82.98 & 80.17 (\worse{-14.62}/\worse{-17.06}) & - & - & - & -\\ %
DCFace \cite{dcface}  & 0.5M           & 98.55 & 85.33 & 82.62 & 89.70 & 91.60 & 89.56 (\worse{-5.23}/\worse{-7.67}) & 64.22 & 62.83 & 42.11 & 49.90  \\ %
IDnet \cite{idnet23} & 0.5M           & 92.58 & 75.40 & 74.25 & 63.88 & 79.90 & 79.13 (\worse{-15.66}/\worse{-18.10}) & - & - & - & -\\ %
Arc2Face \cite{arc2face} & 0.5M           & 98.81 & \underline{91.87} & 85.16 & 90.18 & 92.63 & 91.73 (\worse{-3.06}/\worse{-5.50})  & - & - & - & - \\ 
ID$^3$~\cite{Li2024ID3} & 0.5M & 97.68 & 86.84 & 82.77 & 91.00 & 90.73 & 89.80 (\worse{-4.99}/\worse{-7.43}) & - & - & - & -\\
CeimiFace \cite{cemiface} & 0.5M           & \underline{99.03} & 91.06 & \underline{87.65} & 91.33 & 92.42 & \underline{92.30} (\worse{-2.49}/\worse{-4.93}) & 68.68 & 63.80 & 82.12 & 85.96 \\ 
SFace2 \cite{sface224} & 0.6M           & 95.60 & 77.11 & 74.60 & 77.37 & 83.40 & 81.62 (\worse{-13.17}/\worse{-15.61}) & - & - & - & -\\ %
HyperFace \cite{hyperface} & 0.5M           & 98.50 & 88.83 & 84.23 & 86.53 & 89.40 & 89.50 (\worse{-5.29}/\worse{-7.73}) & - & - & - & -\\
HSFace10K \cite{vec2face25} & 0.5M           & 98.87 & 88.97 & 85.47 & \underline{93.12} & \underline{93.57} & 92.00 (\worse{-2.79}/\worse{-5.23}) & 69.47 & 64.55 & \textbf{83.82} & \textbf{86.96}\\ 
Langevin-Disco \cite{langevin-disco} & 0.6M           & 97.07 & 79.56 & 76.73 & 83.38 & 89.05 & 85.16 (\worse{-9.63}/\worse{-12.07}) & - & - & - & -\\
\textbf{VFace10K (ours)} & 0.5M           & \textbf{99.35} & \textbf{93.56} & \textbf{88.03} & \textbf{94.33} & \textbf{94.17} & \textbf{93.89} (\worse{-0.90}/\worse{-3.34}) & \textbf{70.65} & \textbf{65.63} & \underline{82.92} & \underline{85.75}\\ \hline

DigiFace & 1.2M           & 96.17 & 89.81 & 82.23 & 81.10 & 82.55 & 86.37 (\worse{-8.42}/\worse{-10.86}) & - & - & - & -\\
DCFace & 1.2M           & 98.58 & 88.61 & 85.07 & 90.97 & 92.82 & 91.21 (\worse{-3.58}/\worse{-6.02}) & - & - & - & -\\
Arc2Face  & 1.2M           & 98.92 & \textbf{94.58} & 86.45 & 92.45 & 93.33 & 93.14 (\worse{-1.65}/\worse{-4.09}) & - & - & - & -\\
CemiFace & 1.2M           & 99.22 & 92.84 & \underline{88.86} & 92.13 & 93.03 & \underline{93.22} (\worse{-1.57}/\worse{-4.01}) & - & - & - & -\\
HyperFace & 3.2M           & 98.27 & 92.24 & 85.60 & 90.40 & 91.48 & 91.60 (\worse{-3.19}/\worse{-5.63}) & - & - & - & -\\
HSFace20K & 1M           & 98.87 & 89.87 & 86.13 & \underline{93.85} & \underline{93.65} & 92.47 (\worse{-2.32}/\worse{-4.76}) & 75.22 & 67.55 & 85.31 & 88.78 \\
\textbf{VFace20K (ours)} & 1M           & \textbf{99.40} & \underline{94.23} & \textbf{89.10} & \textbf{95.58} & \textbf{94.67} & \textbf{94.60} (\worse{-0.19}/\worse{-2.63}) & 73.22 & 67.88 & 83.45 & 84.32\\ \hline
HSFace300K  & 15M           & 99.30 & 91.54 & 87.70 & 94.45 & 94.58 & 93.52 (\worse{-1.27}/\worse{-3.71}) & \textbf{81.53} & \textbf{71.32} & 85.18 & \textbf{89.08} \\
\textbf{VFace100K (ours)} & 4M           & 99.52 & \textbf{95.09} & 90.23 & \textbf{95.17} & 94.38 & 94.88 (\better{+0.09}/\worse{-2.35}) & 72.40 & 66.58 & 72.17 & 68.81\\
\textbf{VFace300K (ours)} & 12M           & \textbf{99.57} & 94.80 & \textbf{90.43} & \textbf{95.17} & \textbf{94.70} & \textbf{94.93} (\better{\textbf{+0.14}}/\worse{\textbf{-2.30}}) & 72.37 & 69.38 & \textbf{85.56} & 88.28 \\
\hline \hline
\multicolumn{1}{l|}{CASIA-WebFace} & 0.49M      & 99.38 & 96.91  & 89.78 & 94.50 & 93.35 & \multicolumn{1}{c|}{94.79}  & 77.82 & 68.52 & 78.71 & 83.44 \\ 
\multicolumn{1}{l|}{WebFace4M} & 4M      & 99.73 & 98.93 & 93.93 & 97.52 & 96.02 & \multicolumn{1}{c|}{97.23} & 87.45 & 79.83 & 95.28 & 96.84 \\\hline
\end{tabular}
\label{tab:performance}
\end{table*}

%% file: table/similar-looking.tex
\setlength{\tabcolsep}{1.8mm}
\begin{table}
    \centering
    \caption{Method comparison on similar-looking test sets. Twins-IND and DoppelVer-vise focus on identical twins and doppelgangers. * means reproduced. Numbers equal or less than 50\% are in \textcolor{red}{red}.} %
    \begin{tabular}{l|cc|c}
    \hline
        Datasets & Twins-IND$^\star$ & DoppelVer-vise$^\star$ & Average \\ \hline
        DigiFace & \textcolor{red}{49.47} & 62.17 & 56.09\\
        SFace & \textcolor{red}{50.00} & 77.99 & 64.00\\
        IDiff-Face & \textcolor{red}{50.00} & 81.81 & 65.91\\
        DCFace & \textcolor{red}{49.95} & 88.67 & 69.31\\
        CemiFace & \textcolor{red}{49.75} & 89.54 & 69.65\\
        HSFace10K & \textcolor{red}{49.95} & 86.91 & 68.43\\
        \textbf{VFace10K (ours)} & \textcolor{red}{49.78} & 90.71 & \textbf{70.25} \\ \hline
        HSFace20K & \textcolor{red}{49.97} & 88.90 & 69.43\\
        HSFace300K & \textcolor{red}{49.98} & 90.52 & 70.25\\
        \textbf{VFace20K (ours)} & \textcolor{red}{49.40} & 92.34 & 70.87\\ 
        \textbf{VFace100K (ours)} & \textcolor{red}{50.00} & 92.33 & 71.17\\
        \textbf{VFace300K (ours)} & 50.15 & 93.03 & \textbf{71.59}\\ \hline
        CASIA-WebFace (real) & 54.17 & 95.10 & 74.64\\ 
        WebFace4M (real) & 64.20 & 96.47 & 80.33\\ 
        \hline
    \end{tabular}
    \label{tab:sl-acc}
\end{table}

%% file: table/bias.tex
\setlength{\tabcolsep}{1.8mm}

\begin{table*}[h]
\centering
\caption{Demographic analysis of seven datasets. We observe that models trained by synthetic data have more bias. This table shows true positive rates (TPR@FPR=0.1) of the FR models trained with seven datasets on the BA-test (\textbf{left}) and BFW (\textbf{right}). $\Delta$ calculates gender accuracy difference (Male - Female). For each dataset, the demographic group with the \highest{highest} and \lowest{lowest} accuracy are highlighted. [Keys: Asian (A), Indian (I), Black (B), White (W), Male (M), Female (F), \textcolor{blue}{$\Delta>0$}, \textcolor{red}{$\Delta<0$}, \textcolor{gray}{accuracy $<60$}]}
\label{tab:demog-acc}
\begin{tabular}{c|cc|cc|cc|cc||cc|cc|cc|cc}
\hline
Dataset & AM & AF & IM & IF & BM & BF & WM & WF & AM & AF & IM & IF & BM & BF & WM & WF\\
\hline

DigiFace & \cellcolor[HTML]{EC9494}60.56 & 66.78 & \textcolor{gray}{57.33} & \textcolor{gray}{69.67} & \textcolor{gray}{57.89} & \textcolor{gray}{66.67} & 60.89 & \cellcolor[HTML]{A9D08E}69.67 & \textcolor{gray}{41.79} & \textcolor{gray}{45.33} & \textcolor{gray}{54.24} & \textcolor{gray}{61.38} & \textcolor{gray}{44.09} & \textcolor{gray}{52.05} & \textcolor{gray}{60.43} & \textcolor{gray}{56.94}\\
$\Delta$ &  \multicolumn{2}{c|}{\textcolor{red}{-6.22}}  &  \multicolumn{2}{c|}{\textcolor{gray}{-12.34}}  &  \multicolumn{2}{c|}{\textcolor{gray}{-8.78}}  &  \multicolumn{2}{c||}{\textcolor{red}{-8.87}}&  \multicolumn{2}{c|}{\textcolor{gray}{-3.54}}  &  \multicolumn{2}{c|}{\textcolor{gray}{-7.14}}  &  \multicolumn{2}{c|}{\textcolor{gray}{-7.96}}  &  \multicolumn{2}{c}{\textcolor{gray}{3.49}}\\ \hline

SFace & 77.56 & 74.67 & \cellcolor[HTML]{A9D08E}78.33 & 78.00 & 78.00 & 77.44 & \cellcolor[HTML]{EC9494}69.33 & 73.00 & 69.94 & 70.16 & \cellcolor[HTML]{A9D08E}77.16 & 76.19 & 76.39 & 70.88 & 72.15 & \cellcolor[HTML]{EC9494}69.01\\
$\Delta$ &  \multicolumn{2}{c|}{\textcolor{blue}{2.89}}  &  \multicolumn{2}{c|}{\textcolor{blue}{0.33}}  &  \multicolumn{2}{c|}{\textcolor{blue}{0.56}}  &  \multicolumn{2}{c||}{\textcolor{red}{-3.67}}&  \multicolumn{2}{c|}{\textcolor{red}{-0.22}}  &  \multicolumn{2}{c|}{\textcolor{blue}{0.97}}  &  \multicolumn{2}{c|}{\textcolor{blue}{5.51}}  &  \multicolumn{2}{c}{\textcolor{blue}{3.14}}\\ \hline

IDiff-Face & 86.22 & \cellcolor[HTML]{EC9494}82.67 & 88.33 & 91.22 & 87.22 & 87.89 & 88.89 & \cellcolor[HTML]{A9D08E}91.78 & \cellcolor[HTML]{EC9494}71.97 & 72.37 & 83.45 & \cellcolor[HTML]{A9D08E}84.86 & 82.37 & 78.69 & 80.79 & 80.85\\
$\Delta$ &  \multicolumn{2}{c|}{\textcolor{blue}{3.55}}  &  \multicolumn{2}{c|}{\textcolor{red}{-2.89}}  &  \multicolumn{2}{c|}{\textcolor{red}{-0.67}}  &  \multicolumn{2}{c||}{\textcolor{red}{-2.89}}&  \multicolumn{2}{c|}{\textcolor{red}{-0.4}}  &  \multicolumn{2}{c|}{\textcolor{red}{-1.41}}  &  
\multicolumn{2}{c|}{\textcolor{blue}{3.68}}  &  \multicolumn{2}{c}{\textcolor{red}{-0.06}}\\ \hline

DCFace & 94.33 & 93.11 & 96.11 & \cellcolor[HTML]{A9D08E}98.33 & 96.11 & 96.89 & \cellcolor[HTML]{EC9494}93.33 & 97.22 & 66.60 & 74.89 & 67.97 & \cellcolor[HTML]{A9D08E}84.53 & 67.39 & 74.05 & \cellcolor[HTML]{EC9494}63.37 & 72.35 \\
$\Delta$ &  \multicolumn{2}{c|}{\textcolor{blue}{1.22}}  &  \multicolumn{2}{c|}{\textcolor{red}{-2.22}}  &  \multicolumn{2}{c|}{\textcolor{red}{-0.78}}  &  \multicolumn{2}{c||}{\textcolor{red}{-3.89}}&  \multicolumn{2}{c|}{\textcolor{red}{-8.29}}  &  \multicolumn{2}{c|}{\textcolor{red}{-16.56}}  &  \multicolumn{2}{c|}{\textcolor{red}{-6.66}}  &  \multicolumn{2}{c}{\textcolor{red}{-8.98}}\\ \hline

HSFace10K & 90.33 & \cellcolor[HTML]{EC9494}85.11 & 92.78 & 94.00 & 94.11 & 92.33 & 93.44 & \cellcolor[HTML]{A9D08E}96.22 & \cellcolor[HTML]{EC9494}60.80 & 62.52 & 74.17 & \cellcolor[HTML]{A9D08E}74.64 & 70.76 & 69.52 & 72.36 & 70.19\\
$\Delta$ &  \multicolumn{2}{c|}{\textcolor{blue}{5.22}}  &  \multicolumn{2}{c|}{\textcolor{red}{-1.22}}  &  \multicolumn{2}{c|}{\textcolor{blue}{1.78}}  &  \multicolumn{2}{c||}{\textcolor{red}{-2.78}}&  \multicolumn{2}{c|}{\textcolor{red}{-1.72}}  &  \multicolumn{2}{c|}{\textcolor{red}{-0.47}}  &  \multicolumn{2}{c|}{\textcolor{blue}{1.24}}  &  \multicolumn{2}{c}{\textcolor{blue}{2.17}}\\ \hline

VFace10K & 92.67 & \cellcolor[HTML]{EC9494}82.22 & 93.78 & 94.00 & 94.22 & 93.11 & 93.00 & \cellcolor[HTML]{A9D08E}94.89 & 66.21 & \cellcolor[HTML]{EC9494}64.95 & 76.14 & \cellcolor[HTML]{A9D08E}77.51 & 72.07 & 70.17 & 72.19 & 71.91\\
$\Delta$ &  \multicolumn{2}{c|}{\textcolor{blue}{10.45}}  &  \multicolumn{2}{c|}{\textcolor{red}{-0.22}} & \multicolumn{2}{c|}{\textcolor{blue}{1.11}}  &  \multicolumn{2}{c|}{\textcolor{red}{-1.89}}   &  \multicolumn{2}{c|}{\textcolor{blue}{1.26}} &  \multicolumn{2}{c|}{\textcolor{red}{-1.37}}  &  \multicolumn{2}{c|}{\textcolor{blue}{1.90}}  &  \multicolumn{2}{c}{\textcolor{blue}{0.28}}\\ \hline

CASIA & \cellcolor[HTML]{EC9494}98.78 & 99.33 & 99.11 & \cellcolor[HTML]{A9D08E}99.78 & 99.56 & 98.89 & 99.67 & \cellcolor[HTML]{A9D08E}99.78 & \cellcolor[HTML]{EC9494}89.52 & 91.56 & 92.69 & \cellcolor[HTML]{A9D08E}94.57 & 92.65 & 92.15 & 93.59 & 92.12\\
$\Delta$ &  \multicolumn{2}{c|}{\textcolor{red}{-0.55}}  &  \multicolumn{2}{c|}{\textcolor{red}{-0.67}}  &  \multicolumn{2}{c|}{\textcolor{blue}{0.67}}  &  \multicolumn{2}{c||}{\textcolor{red}{-0.11}}&  \multicolumn{2}{c|}{\textcolor{red}{-2.04}}  &  \multicolumn{2}{c|}{\textcolor{red}{-1.88}}  &  \multicolumn{2}{c|}{\textcolor{blue}{0.5}}  &  \multicolumn{2}{c}{\textcolor{blue}{1.47}}\\ 
\hline
\end{tabular}
\end{table*}

%% file: table/ablation.tex
\setlength{\tabcolsep}{1.0mm}
\begin{table}[!t]
    \centering
    \caption{Ablation study of each processing step for dataset assembling. It indicates that the images generated by AttrOP benefit the model performance on age-oriented test sets and the pose control method can increase the accuracy on pose-oriented test sets. Combing both obtains all the benefits.}
    \label{tab:ablation}
    \begin{tabular}{l|ccccc|c}
    \hline
        VFace10K &  LFW   & CFP-FP & CPLFW & AgeDB & CALFW & Avg. \\ \hline
        Random & 98.66 & 82.17 & 80.25 & 91.45 & 92.67 & 91.04\\
        + AttrOP & 99.33 & 92.91 & 87.88 & 94.33 & 93.82 & 93.66\\
        + pose control & 99.17 & 93.20 & 87.78 & 93.20 & 93.83 & 93.59 \\
        + combined & \textbf{99.35} & \textbf{93.56} & \textbf{88.03} & \textbf{94.33} & \textbf{94.17} & \textbf{93.89}\\
    \hline
    \end{tabular}
\end{table}

%% file: figures/correlation-acc-inter.tex
\begin{figure*}[t]
    \centering
    \includegraphics[width=0.85\linewidth]{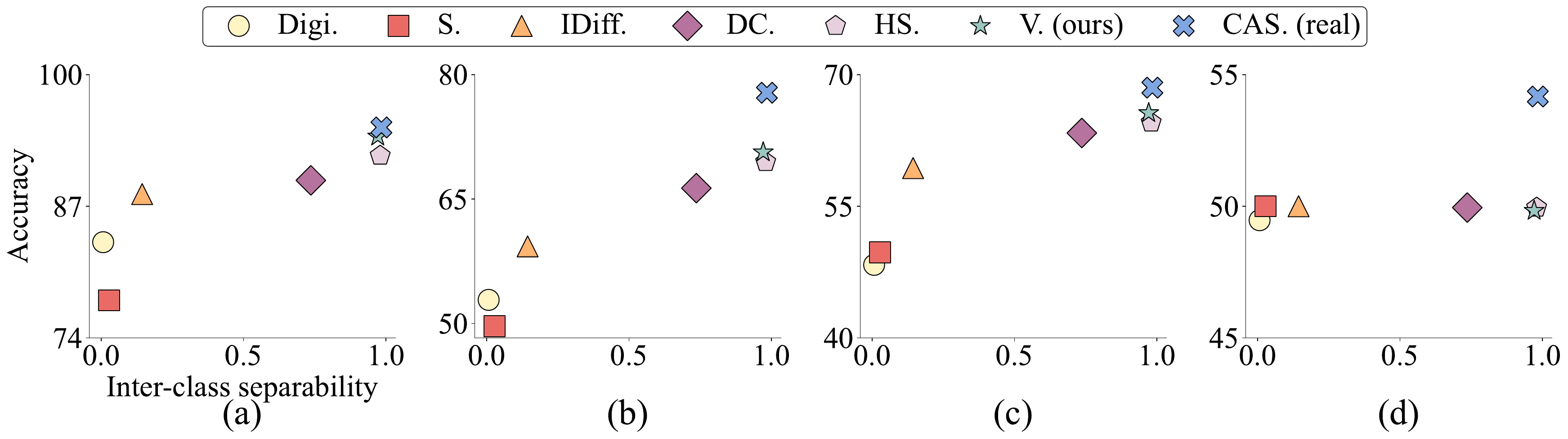}
    \caption{Correlation between inter-class separability and test accuracy. From left to right: test on (a) five test sets, including LFW, CFP-FP, CPLFW, AgeDB-30, and CALFW, (b) Hadrian, (c) Eclipse, and (d) Twins-IND. %
    Separability is calculated using a pre-trained FR model (Section~\ref{sec:inter-calc}). VFace10K and HSFace10K share similarly high inter-class separability. } %
    \label{fig:corr-sep}
\end{figure*}

%% file: figures/correlation-acc-intra.tex
\begin{figure*}[h]
    \centering
    \includegraphics[width=0.85\linewidth]{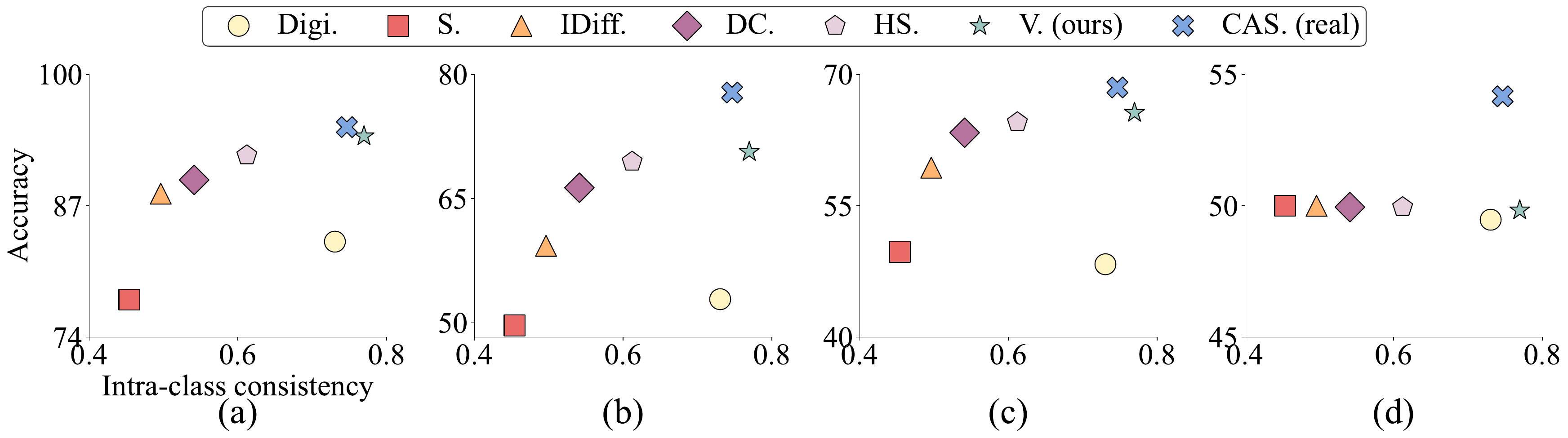}
    \caption{Correlation between intra-class identity consistency and test accuracy. From left to right: test on (a) five test sets, including LFW, CFP-FP, CPLFW, AgeDB-30, and CALFW, (b) Hadrian, (c) Eclipse, and (d) Twins-IND. Consistency is computed according to Section~\ref{sec:intra-calc}. VFace has higher consistency than other synthetic datasets and higher accuracy. }
    \label{fig:corr-consis}
\end{figure*}

%% file: figures/correlation-acc-inter-human.tex
\begin{figure*}[t]
    \centering
    \includegraphics[width=0.85\linewidth]{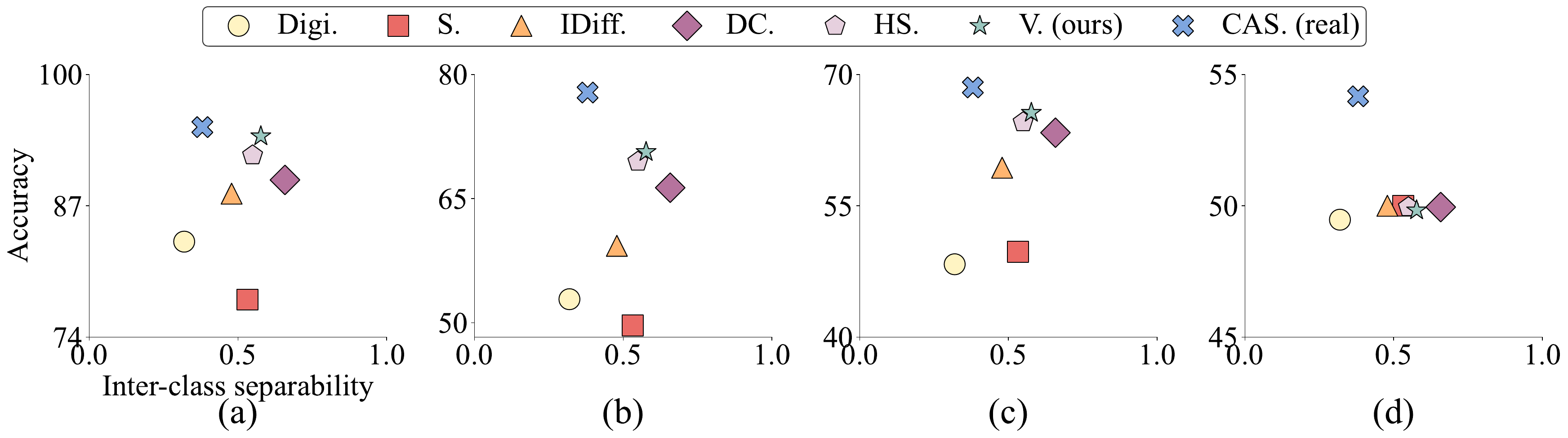}
    \caption{Inter-class separability measured by human vs. test accuracy. %
    We ask human annotators to work on the most similar inter-class pairs in each dataset and tell whether they are of the same identity. Since we selected those most similar inter-class image pairs, this result indicates that inter-class identity separability does not affect a lot on accuracy after a certain degree. It is consistent to Fig.~\ref{fig:corr-sep}}
    \label{fig:corr-sep-human}
\end{figure*}

%% file: figures/correlation-acc-intra-human.tex
\begin{figure*}[h]
    \centering
    \includegraphics[width=0.85\linewidth]{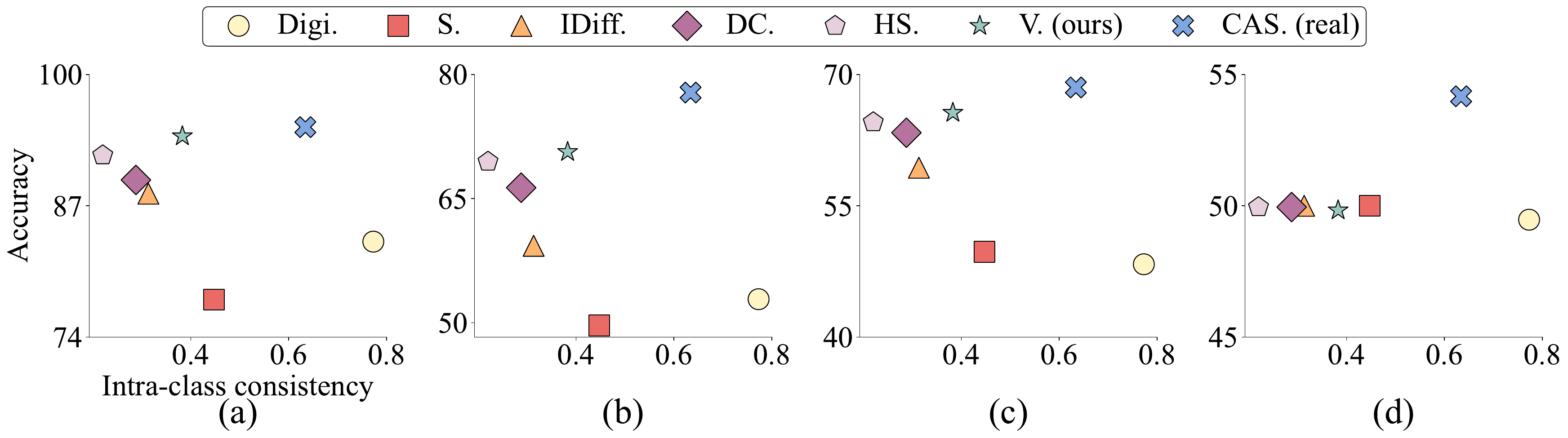}
    \caption{Intra-class ID consistency measured by human vs. test accuracy. %
    We ask human annotators to work on the most dissimilar intra-class pairs in each dataset and tell whether they are of the same identity. Except DigiFace, the other datasets have a worse identity consistency than the real dataset. This suggests that intra-class identity noise can be the cause of the accuracy gap.}%
    \label{fig:corr-consis-human}
\end{figure*}